\RequirePackage{fix-cm}
\documentclass[twocolumn]{svjour3}

\usepackage{natbib}
\usepackage{graphicx}%
\usepackage{multirow}%
\usepackage{amsmath,amssymb,amsfonts}%
\usepackage{mathrsfs}%
\usepackage[title]{appendix}%
\usepackage{xcolor}%
\usepackage{textcomp}%
\usepackage{manyfoot}%
\usepackage{booktabs}%
\usepackage{algorithm}%
\usepackage{algorithmicx}%
\usepackage{algpseudocode}%
\usepackage{listings}%
\usepackage{colortbl}
\usepackage[USenglish]{babel}
\usepackage{hyperref} 
\hypersetup{breaklinks=true,colorlinks,citecolor=black}

\graphicspath{{./Figures/}}

\begin{document}
\title{FADE: A Task-Agnostic Upsampling Operator for Encoder-Decoder Architectures\thanks{\textit{Corresponding author: Zhiguo Cao.
			}			
		}
	}

	\author{Hao Lu$^{1}$ \and  Wenze Liu$^{1}$ \and Hongtao Fu$^{1}$ \and Zhiguo Cao$^{1}$
	}

	\institute{
            Hao Lu$^{1}$ \at
		\email{hlu@hust.edu.cn}         		
		\and
		Wenze Liu$^{1}$ \at
		\email{wzliu@hust.edu.cn}
		\and
		Hongtao Fu$^{1}$ \at
		\email{htfu@hust.edu.cn}
		\and
		Zhiguo Cao$^{1}$ \at
		\email{zgcao@hust.edu.cn}
		\\	
		\and $^{1}$	The Key Laboratory of Image Processing and Intelligent Control, Ministry of Education; School of Artificial Intelligence and Automation, Huazhong University of Science and Technology, Wuhan 430074, China\\
	}

	\date{Received: date / Accepted: date}

	\maketitle





\begin{abstract}{
The goal of this work is to develop a task-agnostic feature upsampling operator for dense prediction where the operator is required to facilitate not only region-sensitive tasks like semantic segmentation but also detail-sensitive tasks such as image matting. Prior upsampling operators often can work well in either type of the tasks, but not both. We argue that task-agnostic upsampling should dynamically trade off between semantic preservation and detail delineation, instead of having a bias between the two properties. In this paper, we present FADE, a novel, plug-and-play, lightweight, and task-agnostic upsampling operator by fusing the assets of decoder and encoder features at three levels: i) considering both the encoder and decoder feature in upsampling kernel generation; ii) controlling the per-point contribution of the encoder/decoder feature in upsampling kernels with an efficient semi-shift convolutional operator; and iii) enabling the selective pass of encoder features with a decoder-dependent gating mechanism for compensating details. To improve the practicality of FADE, we additionally study parameter- and memory-efficient implementations of semi-shift convolution. We analyze the upsampling behavior of FADE on toy data and show through large-scale experiments that FADE is task-agnostic with consistent performance improvement on a number of dense prediction tasks with little extra cost. For the first time, we demonstrate robust feature upsampling on both region- and detail-sensitive tasks successfully. Code is made available at: \url{https://github.com/poppinace/fade}
}


\end{abstract}


\maketitle

\section{Introduction}\label{sec:introduction}

Feature quality, being an important yet hard-to-quantify indicator, significantly influences the performance of a vision system~\citep{girshick2014rich}. This is particularly true for dense prediction tasks such as semantic segmentation~\citep{long2015fully} and object detection~\citep{ren2015faster}, where the predictions highly correlate with the responses of feature maps \citep{zhou2016learning}. Prior art has proposed various ways to enhance the feature quality by operating features, including, but not limited to, spatial pooling \citep{chen18v3,zhao2017pyramid}, feature pyramid fusion \citep{lin2017feature,liu2018path}, attention manipulation \citep{wang2018non}, context aggregation
 \citep{yuan2021ocnet}, and feature alignment \citep{li2020semantic,huang2021fapn}. Yet, the most famous segmentation model \citep{Kirillov_2023_ICCV} so far still struggles to generate accurate boundary predictions, which suggests feature quality remains unsatisfactory. In this work, we delve into an easily overlooked yet fundamental component that closely relates to feature quality---feature upsampling.

Feature upsampling, which aims to recover the spatial resolution of features, is an indispensable stage in most dense prediction models \citep{ronneberger2015u,badrinarayanan2017segnet,xiao2018unified,wang2020deep,zheng2021rethinking,xie2021segformer} as almost all dense prediction tasks prefer high-res predictions. Since feature upsampling is often close to the prediction head, the quality of upsampled features can provide a direct implication of the prediction quality. A good upsampling operator would therefore contribute to improved feature quality and prediction. Yet, conventional upsampling operators, such as nearest neighbor (NN) or bilinear interpolation \citep{lin2017refine}, deconvolution \citep{zeiler2014visualizing}, max unpooling \citep{badrinarayanan2017segnet}, and pixel shuffle \citep{shi2016real}, often have a preference of a specific task. For instance, bilinear interpolation is favored in semantic segmentation \citep{chen18v3,xie2021segformer}, while pixel shuffle is preferred in image super-resolution \citep{Ignatov_2021_CVPR}. 

A main reason is that each dense prediction task has its own focus: some tasks like semantic segmentation \citep{long2015fully} and instance segmentation \citep{he2017mask} are region-sensitive, while some tasks such as image super-resolution \citep{dong2015image} and image matting \citep{xu2017deep,lu2019indices} are detail-sensitive. If one expects an upsampling operator to generate semantically consistent features such that a region can share the same class label, it is often difficult for the same operator to recover boundary details simultaneously, and vice versa. Indeed empirical evidence shows that bilinear interpolation and max unpooling have inverse behaviors in segmentation and matting \citep{lu2019indices,lu2022index}, respectively.

\begin{figure}[!t]
	\centering
	\includegraphics[width=\linewidth]{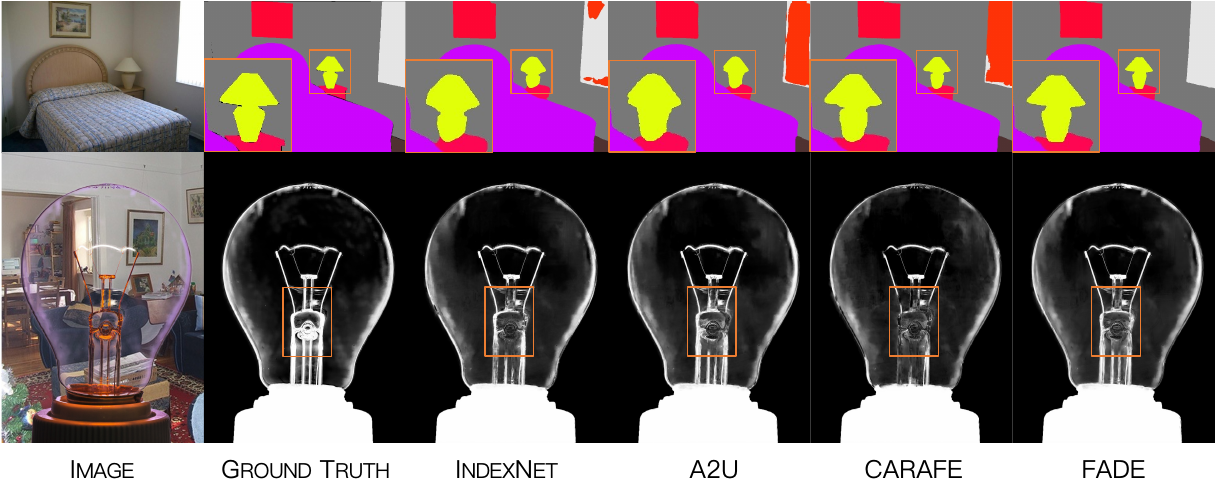}
	\caption{\textbf{Inferred segmentation masks and alpha mattes with different upsampling operators.} The compared operators include IndexNet \citep{lu2019indices}, A2U \citep{dai2021learning}, CARAFE \citep{jiaqi2019carafe}, and our proposed FADE. Among competitors, only FADE generates both the high-quality mask and the alpha matte.}
	\label{fig:seg_matte_intro}
\end{figure}

In an effort to evade `trials-and-errors' from choosing an upsampling operator for a certain task at hand, there has been a growing interest in developing a generic upsampling operator for dense prediction \citep{mazzini2018guided,tian2019decoders,jiaqi2019carafe,wang2020carafe++,lu2019indices,lu2022index,dai2021learning}. For example, CARAFE \citep{jiaqi2019carafe} shows its benefits on four dense prediction tasks, including object detection, instance segmentation, semantic segmentation, and image inpainting. IndexNet \citep{lu2019indices} also boosts performance on several tasks such as image matting, image denoising, depth prediction, and image reconstruction. However, a comparison between CARAFE and IndexNet \citep{lu2022index} indicates that neither CARAFE nor IndexNet can defeat its opponent on both region- and detail-sensitive tasks (CARAFE outperforms IndexNet on segmentation, while IndexNet can surpass CARAFE on matting), which can also be observed from the inferred segmentation masks and alpha mattes in Fig.~\ref{fig:seg_matte_intro}. This raises a fundamental research question: \textit{What makes for task-agnostic upsampling?} 

\begin{figure*}[!t]
	\centering
	\includegraphics[width=\linewidth]{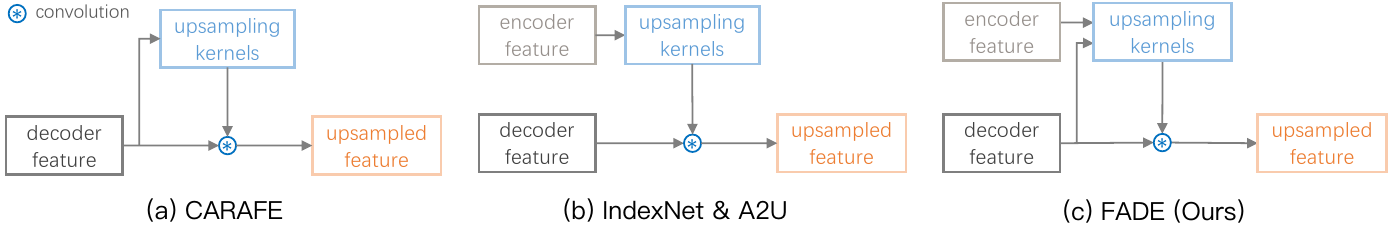}
	\caption{\textbf{Main difference between dynamic upsampling operators on the use of encoder and/or decoder features.} (a) CARAFE \citep{jiaqi2019carafe} generates upsampling kernels conditioned on decoder features, while (b) IndexNet \citep{lu2022index} and A2U \citep{dai2021learning} generate kernels using encoder features only. By contrast, (c) FADE considers both encoder and decoder features in upsampling kernel generation.}
	\label{fig:concept_map}
\end{figure*}

After an apples-to-apples comparison between existing dynamic upsampling operators (Fig.~\ref{fig:concept_map}), we hypothesize that it is the inappropriate and/or insufficient use of high-res encoder and low-res decoder features that leads to the task dependency of upsampling. We also believe that there should exist a unified form of upsampling operator that is truly task-agnostic. In particular, we argue that \textit{a task-agnostic upsampling operator should dynamically trade off between semantic preservation and detail delineation in a content-aware manner, instead of having a bias between the two properties}. To this end, our main idea is to make the full use of encoder and decoder features in upsampling (kernels). We therefore introduce FADE, a novel, plug-and-play, lightweight, and task-agnostic upsampling operator for encoder-decoder architectures. The name also implies its working mechanism: upsampling features in a `fade-in' manner, from recovering spatial structure to delineating subtle details.
In the context of hierarchical encoder-decoder architectures such as feature pyramid networks (FPNs) \citep{lin2017feature} and U-Net \citep{ronneberger2015u}, semantic information is 
rich in low-res decoder features, and detailed information is often abundant in high-res encoder features. To exploit both information in feature upsampling, FADE Fuses the Assets of Decoder and Encoder with three key observations and designs:
\begin{enumerate}

    \item[i)] By exploring why CARAFE works well on region-sensitive tasks but poorly on detail-sensitive tasks, and why IndexNet and A2U \citep{dai2021learning} behave conversely, we observe that what features (encoder or decoder) to use to generate the upsampling kernels matters. Using low-res decoder features preserves regional coherence, while using high-res encoder features helps recover details. It is thus natural to seek whether combining encoder and decoder features enjoys both merits, which underpins the core idea of FADE, as shown in Fig.~\ref{fig:concept_map}.
    
    \item[ii)] To integrate high-res encoder and low-res decoder features, a subsequent obstacle is how to deal with the problem of resolution mismatch. 
    A standard way is to implement U-Net-style fusion \citep{ronneberger2015u}, including feature interpolation, feature concatenation, and convolution. However, we show that this naive implementation can introduce artifacts into upsampling kernels. To solve this, we introduce a semi-shift convolutional operator that unifies channel compression, concatenation, and kernel generation. Particularly, it allows granular control over how each feature point contributes to upsampling kernels. 

    \item[iii)] Inspired by the gating mechanism used in FPN-like designs \citep{li2020gated,li2023sfnet}, we further refine upsampled features by enabling selective pass of high-res encoder features via a simple decoder-dependent gating unit.
    
\end{enumerate}
To improve the practicality and efficiency of FADE, we also investigate parameter-efficient and memory-efficient implementations of semi-shift convolution. Such implementations lead to a lightweight variant of FADE termed FADE-Lite. We show that, even with one forth number of parameters of FADE, FADE-Lite still preserves the task-agnostic property and behaves reasonably well across different tasks. The memory-efficient implementation also enables direct execution of cross-resolution convolution, without explicit feature interpolation for resolution matching.

We conduct experiments on seven data sets covering six dense prediction tasks. We first validate our motivation and the rationale of our design via several toy-level and small-scale experiments, such as binary image segmentation on Weizmann Horse \citep{borenstein2002class}, image reconstruction on Fashion-MNIST \citep{xiao2017fashion}, and semantic segmentation on SUN RGBD \citep{song2015sun}. We then show through large-scale evaluations that FADE reveals its task-agnostic property by consistently boosting both region- and detail-sensitive tasks, for instance: i) \textit{semantic segmentation}: FADE improves SegFormer-B1 \citep{xie2021segformer} by $+2.73$ mask IoU and $+4.85$ boundary IoU on ADE20K \citep{zhou2017scene} and steadily boosts the boundary IoU with stronger backbones, ii) \textit{image matting}: FADE outperforms the previous best matting-specific upsampling operator A2U \citep{dai2021learning} on Adobe Composition-1K \citep{xu2017deep}, iii) \textit{object detection} and iv) \textit{instance segmentation}: FADE performs comparably against the best performing operator CARAFE over Faster R-CNN \citep{ren2015faster} ($+1.1$ AP for FADE vs.\ $+1.2$ AP for CARAFE with ResNet-50) and Mask R-CNN \citep{he2017mask} ($+0.4$ mask AP for FADE vs.\ $+0.7$ mask AP for CARAFE with ResNet-50) baselines on Microsoft COCO \citep{lin2014microsoft}, and v) \textit{monocular depth estimation}: FADE also surpasses the previous best upsampling operator IndexNet \citep{lu2022index} over the BTS \citep{lee2019big} baseline on NYU Depth V2 \citep{silberman2012indoor}. In addition, FADE retains the lightweight property by introducing only a few amount of parameters and FLOPs. It has also good generality across convolutional and transformer architectures \citep{xiao2018unified,xie2021segformer}.

Overall, our contributions include the following:
\begin{itemize}
    \item For the first time, we show that task-agnostic upsampling is made possible on both high-level region-sensitive and low-level detail-sensitive tasks;

    \item We present FADE, one of the first task-agnostic upsampling operator, that fuses encoder and decoder features in generating upsampling kernels, uses an efficient semi-shift convolutional operator to control per-point contribution, and optionally applies a gating mechanism to compensate details;

    \item We provide a comprehensive benchmarking on state-of-the-art upsampling operators across five mainstream dense prediction tasks, which facilitates future study.
\end{itemize}

A preliminary conference version of this work appeared in \citep{lu2022fade}. We extend \citep{lu2022fade} from the following aspects: i) to highlight the task-agnostic property, we validate FADE comprehensively on more baseline models, \textit{e.g.}, UPerNet \citep{xiao2018unified}, Faster RCNN \citep{ren2015faster}, Mask RCNN \citep{he2017mask}, and BTS \citep{lee2019big}, on different network scales, from SegFormer-B1 to -B5 \citep{xie2021segformer} and from R50 to R101 \citep{he2016deep}, and on three additional vision tasks including object detection, instance segmentation, and monocular depth estimation; ii) we carefully benchmark the performance of state-of-the-art dynamic upsampling operators on the evaluated tasks to provide a basis for future studies; iii) we further explore parameter-efficient and memory-efficient implementations of semi-shift convolution to enhance the practicality of FADE, which also leads to a lightweight variant called FADE-Lite; iv) by observing some unexpected phenomena in experiments, we rethink the value of the gating mechanism in FADE and provide additional analyses and insights on when to use the gating unit, particularly for instance-level tasks; v) we extend the related work by comparing feature upsampling with other closely related techniques such as feature alignment and boundary processing; vi) we also extend our discussion on the general value of feature upsampling to dense prediction.

\section{Literature Review}

We review upsampling operators in deep networks, techniques that share a similar spirit to upsampling including feature alignment and boundary processing, and typical dense prediction tasks in vision.

\subsection{Feature Upsampling} 

Unlike joint image upsampling \citep{tomasi1998bilateral,he2010guided}, feature upsampling operators are mostly developed in the era of deep learning, to respond to the need for recovering spatial resolution of encoder features (decoding). Conventional upsampling operators typically use fixed/hand-crafted kernels. For instance, the kernels in the widely used NN and bilinear interpolation are defined by the relative distance between pixels. Deconvolution \citep{zeiler2014visualizing}, \textit{a.k.a.}\ transposed convolution, also applies a fixed kernel during inference, despite the kernel parameters are learned. Pixel Shuffle \citep{shi2016real} first employs convolution to adjust feature channels and then reduces the depth dimension to increase the spatial  dimension. While the main purpose of resolution increase is achieved, the operators above also introduce certain artifacts into features. For instance, it is well-known that, interpolation smooths boundaries, and deconvolution generates checkerboard artifacts \citep{odena2016deconvolution}. Several recent work has shown that unlearned upsampling has become a bottleneck behind architectural design \citep{liu2023devil}, and dynamic upsampling behaviors are more expected \citep{lu2019indices}. Among hand-crafted operators, unpooling \citep{badrinarayanan2017segnet} perhaps is the only operator that implements dynamic upsampling, \textit{i.e.}, each upsampled position is data-dependent conditioned on the $\max$ operator. The importance of such a dynamic property has been exemplified by some recent dynamic kernel-based upsampling operators \citep{jiaqi2019carafe,lu2019indices,dai2021learning,lu2022sapa}, which leads to a new direction from considering generic feature upsampling across tasks and architectures. In particular, CARAFE \citep{jiaqi2019carafe} implements context-aware reassembly of features with decoder-dependent upsampling kernels, IndexNet \citep{lu2019indices} provides an indexing perspective of upsampling and executes upsampling by learning a soft index (kernel) function, and A2U \citep{dai2021learning} introduces affinity-aware upsampling kernels by exploiting second-order information. At the core of these operators is the data-dependent upsampling kernels whose kernel parameters are not learned but dynamically predicted by a sub-network.

However, while being dynamic, CARAFE, A2U, and IndexNet still exhibit a certain degree of bias on specific tasks. In this work, we show through FADE that the devil is in the use of encoder and decoder features in generating upsampling kernels.

\subsection{Feature Alignment and Boundary Processing}
Different from dynamic upsampling that aims to enhance feature quality \textit{during} resolution change, much existing work also attempts to enhance the feature quality \textit{after} matching resolution.
Two closely related techniques are feature alignment and boundary processing. Feature alignment explores to align multi-level feature maps by warping features with, for example, either sampling offsets \citep{wu2022fsanet,huang2021fapn} or a dense flow field \citep{li2020semantic,li2023sfnet}, which has been found effective in reducing semantic aliasing during cross-resolution feature fusion. Another idea is to use a gating unit to align and refine features \citep{li2020gated}, which prevents encoder noise from entering decoder feature maps. FADE has also a similar design as post-processing, but is much simpler. 
Considering that, most fragile predictions in segmentation are along object boundaries, 
boundary processing techniques are developed to optimize boundary quality. In particular, PointRend \citep{kirillov2020pointrend} views segmentation as a rendering problem and adaptively selects points to predict crisp boundaries by an iterative subdivision algorithm. \citet{li2020improving} improves boundary prediction with decoupled body and edge supervision. Boundary-preserving Mask R-CNN \citep{cheng2020boundary} presents a boundary-preserving mask head to improve mask localization accuracy. Gated-SCNN \citep{takikawa2019gated} introduces a two-stream architecture that wires shape information as a separate processing branch to process boundary-related information specifically.

Compared with dynamic upsampling, feature alignment and boundary processing are typically executed \textit{after} naive feature upsampling. Since feature upsampling is inevitable, it would be interesting to see whether one could enhance the feature quality \textit{during} upsampling, which is exactly one of the goals of dynamic upsampling. In this work, we show that FADE is capable of mitigating semantic aliasing as feature alignment and of improving boundary predictions as boundary processing. FADE also demonstrates universality across a number of tasks more than segmentation.


\subsection{Dense Prediction} 

Dense prediction covers a broad class of per-pixel labeling tasks, ranging from mainstream object detection \citep{ren2015faster}, semantic segmentation \citep{long2015fully}, instance segmentation \citep{he2017mask}, and depth estimation \citep{eigen2014depth} to low-level image restoration \citep{mao2016image}, image matting \citep{xu2017deep}, edge detection \citep{xie2015holistically}, and optical flow estimation \citep{teed2020raft}, to name a few. An interesting property about dense prediction is that a task could be region-sensitive or detail-sensitive. The sensitivity is closely related to what metric is used to assess the task. In this sense, semantic/instance segmentation is region-sensitive, because the standard Mask Intersection-over-Union (IoU) metric \citep{everingham2010pascal} is mostly affected by regional mask prediction quality, instead of boundary quality. On the contrary, image matting can be considered detail-sensitive, because the error metrics \citep{rhemann2009perceptually} are mainly computed from trimap regions that are full of subtle details or transparency. Note that, when we emphasize region sensitivity, we do not mean that details are not important, and vice versa. In fact, the emergence of the Boundary IoU metric \citep{cheng2021boundary} implies that the limitation of a certain evaluation metric has been noticed by our community. 

Feature upsampling can play important roles in dense prediction, not only for generating high-resolution predictions but also for improving the quality of predictions. The goal of developing a task-agnostic and content-aware upsampling operator capable of both regional preservation and detail delineation can have a broad impact on a number of dense prediction tasks. In this work, we evaluate FADE and other upsampling operators on both types of tasks using both region-aware and detail-aware metrics.

\section{Task-Agnostic Upsampling: A Trade-off Between Semantic Preservation and Detail Delineation?}
\label{sec:insights}

Before we present FADE, we share some of our view points towards task-agnostic upsampling, which may be helpful to understand our designs in FADE.

\begin{remark}
Encoder and decoder features play different roles in upsampling, particularly in the generation of upsampling kernels.
\end{remark}

In dense prediction models, downsampling stages are involved to reduce computational burden or to acquire a large receptive field, bringing the need of peer-to-peer upsampling stages to recover the spatial resolution, which together constitutes the basic encoder-decoder architecture. During downsampling, details of high-res features are impaired or even lost, but the resulting low-res encoder features often have good semantic meanings that can pass to decoder features.
Hence, we believe an ideal upsampling operator should appropriately resolve two issues: 1) preserve the semantic information already extracted; 2) compensate as many lost details as possible without deteriorating the semantic information. NN or bilinear interpolation only meets the former. This conforms to our intuition that interpolation often smooths features. A reason is that low-res decoder features have no prior knowledge about missing details. Other operators that directly upsample decoder features, such as deconvolution and pixel shuffle, can have the same problem with poor detail compensation. Compensating details requires high-res encoder features. This is why unpooling that stores indices before downsampling has good boundary delineation \citep{lu2019indices}, but it hurts the semantic information due to zero-filling.

Dynamic upsampling operators, including CARAFE \citep{jiaqi2019carafe}, IndexNet \citep{lu2019indices}, and A2U \citep{dai2021learning}, alleviate the problems above with data-dependent upsampling kernels. Their upsampling modes are shown in Fig.~\ref{fig:concept_map}(a)-(b). From Fig.~\ref{fig:concept_map}, it can be observed that, CARAFE generates upsampling kernels conditioned on decoder features, while IndexNet \citep{lu2019indices} and A2U \citep{dai2021learning} generate kernels via encoder features. This may explain the inverse behavior between CARAFE and IndexNet/A2U on region- or detail-sensitive tasks \citep{lu2022index}. Yet, we find that generating upsampling kernels using either encoder or decoder features can lead to suboptimal results, and it is critical \textit{to leverage both encoder and decoder features for task-agnostic upsampling}, as implemented in FADE (Fig.~\ref{fig:concept_map}(c)).

\begin{figure}[!t]
	\centering
	\includegraphics[width=\linewidth]{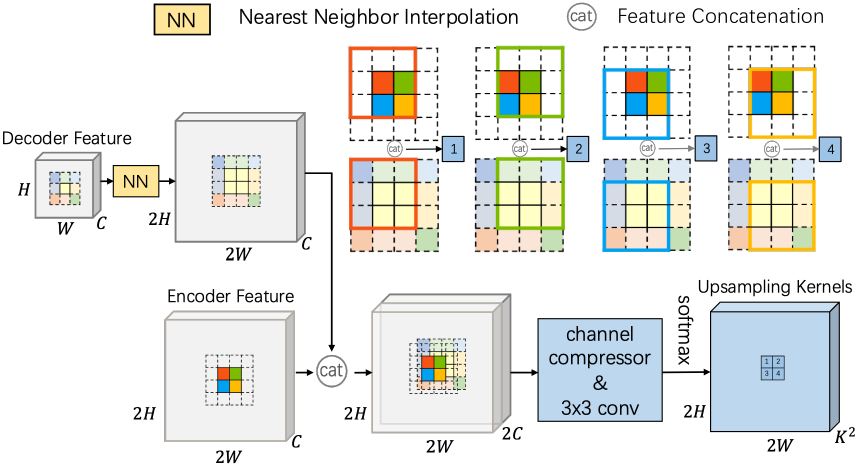}
	\caption{\textbf{Naive implementation for generating upsampling kernels using encoder and decoder features.} The kernel prediction using high-res encoder and low-res decoder features requires matching resolution with explicit feature interpolation and concatenation, followed by channel compression and convolution.}
	\label{fig:unet-style_conv}
\end{figure}

\begin{figure*}[!t]
	\centering
	\includegraphics[width=\linewidth]{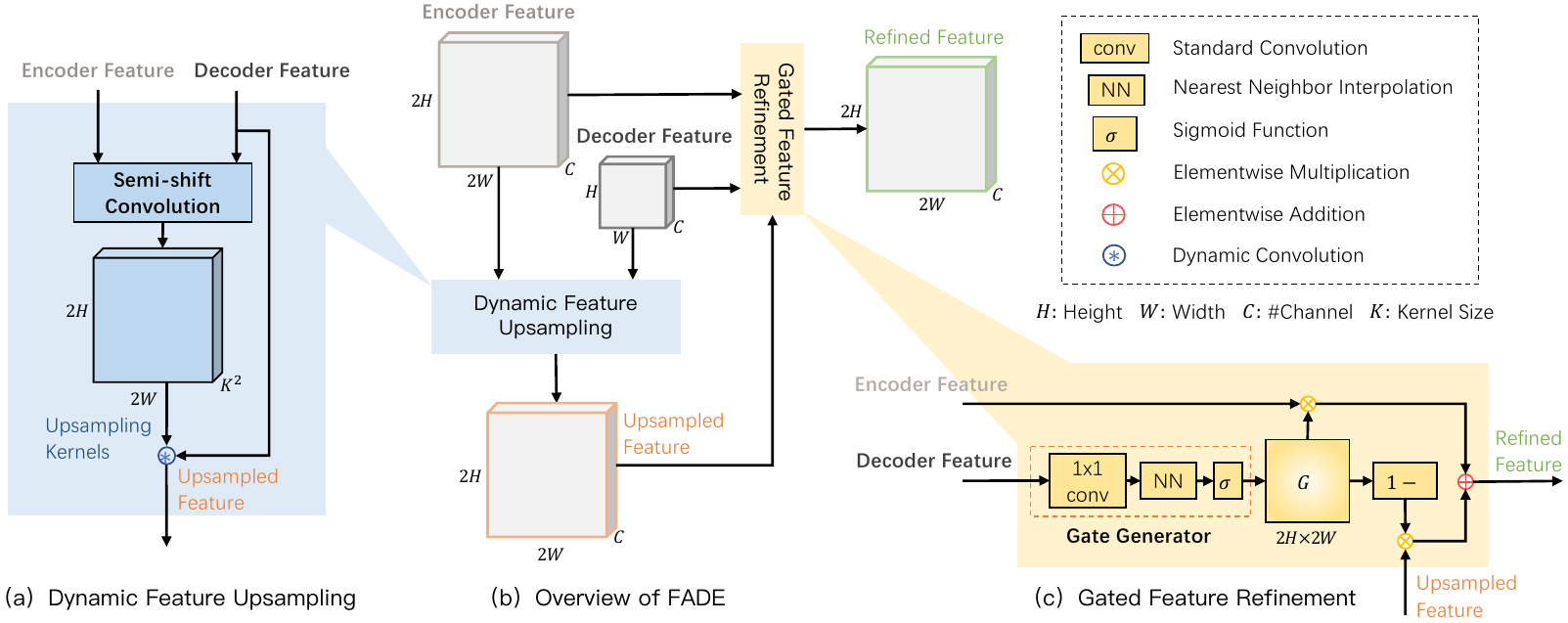}
	\caption{\textbf{Technical pipeline of FADE.} From (b) the overview of FADE, 
	FADE upsamples the low-res decoder feature with the help of the high-res encoder features. The two types of features are fed into two key modules. In (a) dynamic feature upsampling, the features are used to generate upsampling kernels using a semi-shift convolutional operator (Fig.~\ref{fig:semi-shift_conv}). The kernels are then applied to the decoder feature to generate the upsampled feature. In (c) gated feature refinement, the encoder and upsampled features are modulated by a decoder-dependent gating mechanism to enhance detail delineation before outputting the final refined feature.}
	\label{fig:pipeline}
\end{figure*}

\begin{remark}
How each feature point contributes to upsampling matters.
\end{remark}

After deciding what the features to use, the follow-up question is how to use the features effectively and efficiently. The main obstacle is the mismatched resolution between encoder and decoder features. Per Fig.~\ref{fig:unet-style_conv}, one may consider simple interpolation for resolution matching, but this can lead to sub-optimal upsampling. Considering the case of applying $\times2$ NN interpolation to decoder features, if we use $3\times3$ convolution to generate the upsampling kernel, the effective receptive field of the kernel can reduce to be $<50\%$: before interpolation there are $9$ valid points in a $3\times3$ window, but only $4$ valid points are left after interpolation. Besides this, another more important issue remains. 
Still in the $\times2$ upsampling in Fig.~\ref{fig:unet-style_conv}, the four windows which control the variance of upsampling kernels w.r.t.\ the $2\times2$ neighbors of high resolution are affected by the naive interpolation. Controlling a high-res upsampling kernel map, however, is blind with the low-res decoder feature. It contributes little to 
the variance of the four neighbors. 
A more reasonable choice may be to \textit{let encoder and decoder features cooperate to control the overall upsampling kernel, but let the encoder feature alone control the variance of the four neighbors}. This insight exactly motivates the design of semi-shift convolution (Section \ref{subsec:semi-shift}).

\begin{remark}
High-res encoder features can be leveraged for further detail refinement.
\end{remark}

Besides helping structural recovery via upsampling kernels, there remains 
useful information in encoder features. Since encoder features only go through a few layers of a network, they preserve `fine details' of high resolution. In fact, nearly all dense prediction tasks require fine details, \textit{e.g.}, despite regional prediction dominates in instance segmentation, accurate boundary prediction can significantly boost performance \citep{tang2021look}, not to mention the stronger request of fine details in detail-sensitive tasks. \textit{The demands of fine details in dense prediction need further exploitation of encoder features}. Following existing ideas \citep{cho2014properties,li2020gated,li2023sfnet}, we explore the use of a gating mechanism by leveraging low-res decoder features to guide where the high-res encoder features can pass through. Yet, in some instance-aware tasks, we find that the gate is better left fully open (more discussion can be found in Section~\ref{subsec:gate}).

\section{FADE: Fusing the Assets of Decoder and Encoder}
\label{sec:fade}

Here we elaborate our designs in FADE. We first revisit the framework of dynamic upsampling, then present from three aspects on how to fuse the assets of decoder and encoder features in upsampling, particularly discussing the principle and the efficient implementations of the semi-shift convolution. 

\subsection{Dynamic Upsampling Revisited} Here we review some basic operations in recent dynamic upsampling operators such as CARAFE \citep{jiaqi2019carafe}, IndexNet \citep{lu2019indices}, and A2U \citep{dai2021learning}. Fig.~\ref{fig:concept_map} briefly summarizes their upsampling modes. They share an identical pipeline, \textit{i.e.}, first generating data-dependent upsampling kernels, and then reassembling the decoder features using the kernels. Typical dynamic upsampling kernels are content-aware, but channel-shared, which means each position has a unique upsampling kernel in the spatial dimension, but the same ones are shared in the channel dimension.

CARAFE learns upsampling kernels directly from decoder features and then reassembles them to high resolution. Specifically, the decoder features pass through two consecutive convolutional layers to generate the upsampling kernels, of which the former is a channel compressor implemented by $1\times1$ convolution used to reduce the computational complexity and the latter is a content encoder with $3\times3$ convolution. IndexNet and A2U, however, 
adopt more sophisticated modules to leverage the merit of encoder features. Further details can be referred to \citep{jiaqi2019carafe,lu2019indices,dai2021learning}.

FADE is designed to maintain the simplicity of dynamic upsampling. 
Hence, we mainly 
optimize the process of kernel generation with semi-shift convolution, and the channel compressor will also function as a way of pre-fusing encoder and decoder features. In addition, FADE also includes a gating mechanism for detail refinement. The overall pipeline of FADE is summarized in Fig.~\ref{fig:pipeline}. In what follows, we explain our three key designs and present our efficient implementations. 

\begin{table}[!t]\scriptsize
    \caption{Results of semantic segmentation on SUN RGBD and image reconstruction on Fashion MNIST. Best performance is in \textbf{boldface}}
    \label{tab:seg_and_rec}
    \centering
    \renewcommand{\arraystretch}{1.2}
    \addtolength{\tabcolsep}{-2pt}
    \begin{tabular}{@{}lcc|cccc@{}}
    \toprule
    & \multicolumn{2}{c|}{$\tt Segmentation$}  & \multicolumn{4}{c}{$\tt Reconstruction$} \\ 
    & \multicolumn{2}{c|}{$\tt accuracy\uparrow$} & \multicolumn{2}{c}{$ \tt accuracy\uparrow$} & \multicolumn{2}{c}{$\tt error\downarrow$}\\
     & mIoU & bIoU & PSNR & SSIM & MAE & MSE\\ 
    \midrule
    decoder-only & 37.00 & 25.61 & 24.35 & 87.19 & 0.0357 & 0.0643 \\
    encoder-only & 36.71 & 27.89 & 32.25 & 97.73 & 0.0157 & 0.0257 \\
    encoder-decoder & \textbf{37.59} & \textbf{28.80} & \textbf{33.83} & \textbf{98.47} & \textbf{0.0122} & \textbf{0.0218} \\ 
    \bottomrule
    \end{tabular}
\end{table}

\subsection{Generating Upsampling Kernels from Encoder and Decoder Features}

\begin{figure}[!t]
	\centering
	\includegraphics[width=\linewidth]{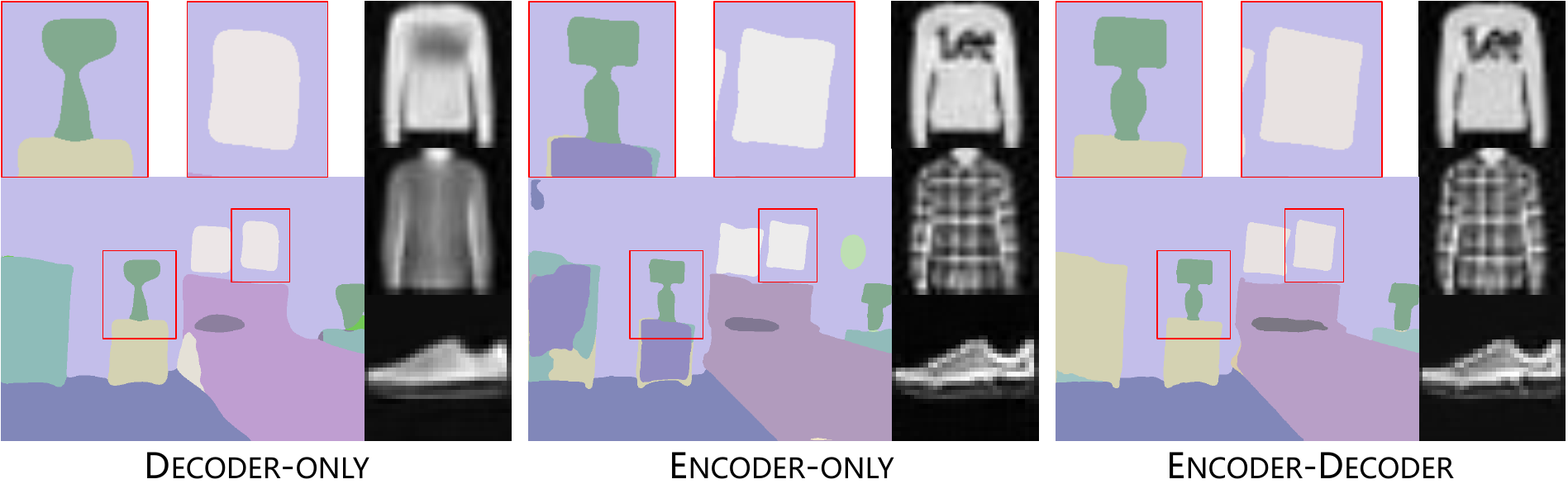}
	\caption{\textbf{Visualizations of inferred mask and reconstructed results on SUN RGBD and Fashion-MNIST.} The decoder-only model generates semantically consistent mask predictions but poor reconstructions, while the encoder-only one is on the contrary. When both encoder and decoder features are considered, the model generates reasonable masks as the decoder-only model and clear reconstructions as the encoder-only one (cf. the table lamp and the stripes on clothes).}
	\label{fig:sunrgbd_visual}
\end{figure}

We first showcase a few visualizations on some small-scale or toy-level data sets to highlight the importance of both encoder and decoder features for task-agnostic upsampling. We choose semantic segmentation on SUN RGBD \citep{song2015sun} as the region-sensitive task and image reconstruction on Fashion MNIST \citep{xiao2017fashion} as the detail-sensitive one. We follow the network architectures and the experimental settings in \citep{lu2022index}. Since we focus on upsampling, all downsampling stages use max pooling. Specifically, to show the impact of encoder and decoder features, in the segmentation experiments, we use CARAFE as the baseline but only modify the source of features used for generating upsampling kernels. We build three baselines: 
1) \textit{decoder-only}, the standard implementation of CARAFE; 
2) \textit{encoder-only}, where the upsampling kernels are generated from encoder features; 
3) \textit{encoder-decoder}, where the upsampling kernels are generated from the concatenation of encoder and NN-interpolated decoder features. 
We report Mask IoU (mIoU) \citep{everingham2010pascal} and Boundary IoU (bIoU) \citep{cheng2021boundary} for segmentation, and Peak Signal-to-Noise Ratio (PSNR), Structural SIMilarity index (SSIM), Mean Absolute Error (MAE), and root Mean Square Error (MSE) for reconstruction. 
From Table~\ref{tab:seg_and_rec}, one can observe that the encoder-only baseline outperforms the decoder-only one in image reconstruction, but in semantic segmentation the trend is on the contrary. To understand why, we visualize the segmentation masks and reconstructed results in Fig.~\ref{fig:sunrgbd_visual}. We find that in segmentation the decoder-only model tends to produce regionally coherent masks, while the encoder-only one generates clear mask boundaries but blocky regions; in reconstruction, by contrast, the decoder-only model almost fails and can only generate low-fidelity reconstructions. It thus can be inferred that, high-res encoder features help to predict details, while low-res decoder features contribute to semantic preservation of regions. Indeed, by considering both encoder and decoder features, the resulting mask seems to integrate the merits of the former two, and the reconstructions are also full of details. Therefore, albeit a simple tweak, FADE significantly benefits from generating upsampling kernels with both encoder and decoder features, as illustrated in Fig.~\ref{fig:concept_map}(c).

\begin{figure}[t]
	\centering
	\includegraphics[width=\linewidth]{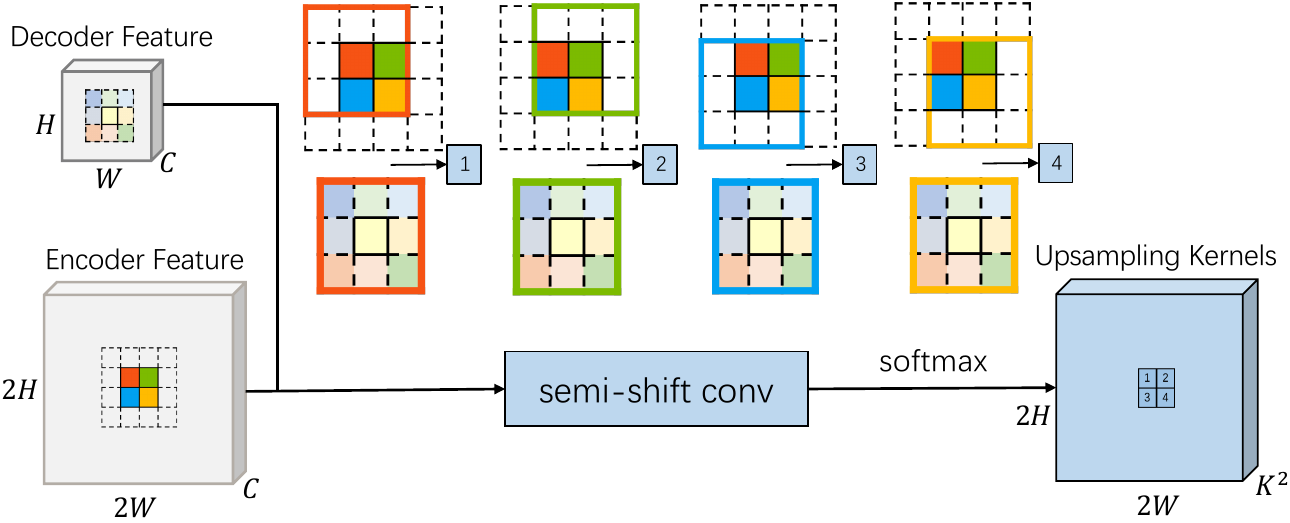}
	\caption{\textbf{Upsampling kernel generation using semi-shift convolution with both encoder and decoder features.} In contrast to naive implementation (Fig.~\ref{fig:unet-style_conv}), semi-shift convolution carefully controls the per-point contribution to the kernel (see how each decoder feature point corresponds to each encoder feature point) and unifies feature interpolation, concatenation, channel compression, and kernel prediction.}
	\label{fig:semi-shift_conv}
\end{figure}

\subsection{Semi-shift Convolution}
\label{subsec:semi-shift}

Given encoder and decoder features, we next address how to use them to generate upsampling kernels. We investigate two implementations: the naive one presented in Fig.~\ref{fig:unet-style_conv} and our customized one -- semi-shift convolution. We first illustrate the principle of semi-shift convolution and then present its efficient implementations. Finally, we compare the computational workload and memory occupation among different implementations.

\subsubsection{Principle of Semi-shift Convolution}

The key difference between naive and semi-shift convolution is how each decoder feature point spatially corresponds to each encoder feature point. The naive implementation shown in Fig.~\ref{fig:unet-style_conv} includes five operations: i) feature interpolation, ii) concatenation, iii) channel compression, iv) standard convolution for kernel generation, and v) {\rm softmax} normalization. As aforementioned in Section \ref{sec:insights}, naive interpolation can have a few problems. To address them, we propose semi-shift convolution that simplifies the first four operations above into a unified operator, which is 
illustrated in Fig.~\ref{fig:semi-shift_conv}. Note that the $4$ convolution windows in encoder features all correspond to the same window in decoder features. This design has the following advantages: 
1) the role of control in the kernel generation is made clear where the control of the variance of $2\times2$ neighbors is moved to encoder features completely;
2) the receptive field of decoder features is kept consistent with that of encoder features;
3) memory cost is reduced, because semi-shift convolution directly operates on low-res decoder features, without feature interpolation;
4) channel compression and kernel generation can also be merged in semi-shift convolution.

Mathematically, the single window processing with naive implementation or semi-shift convolution has an identical form if ignoring the content of feature maps.
For example, considering the top-left window w.r.t. the index `1' in Figures~\ref{fig:unet-style_conv} and~\ref{fig:semi-shift_conv}, the (unnormalized) upsampling kernel takes the form
\begin{equation} \label{eq:1}
\footnotesize
  \begin{split}
      w_m &= \sum\limits_{l=1}^{d}\sum\limits_{i=1}^{h}\sum\limits_{j=1}^{h}\beta_{ijlm}\left(\sum\limits_{k=1}^{2C}\alpha_{kl}x_{ijk} + a_l\right) + b_m \\
      &= \sum\limits_{l=1}^{d}\sum\limits_{i=1}^{h}\sum\limits_{j=1}^{h}\beta_{ijlm}\left(\sum\limits_{k=1}^{C}\alpha_{kl}^{\tt en}x_{ijk}^{\tt en} + \sum\limits_{k=1}^{C}\alpha_{kl}^{\tt de}x_{ijk}^{\tt de} + a_l\right) + b_m \\
      &= \sum\limits_{l=1}^{d}\sum\limits_{i=1}^{h}\sum\limits_{j=1}^{h}\beta_{ijlm}\sum\limits_{k=1}^{C}\alpha_{kl}^{\tt en}x_{ijk}^{\tt en} \\
      & ~~~~~~~~~~~~~ + \sum\limits_{l=1}^{d}\sum\limits_{i=1}^{h}\sum\limits_{j=1}^{h}\beta_{ijlm}\left(\sum\limits_{k=1}^{C}\alpha_{kl}^{\tt de}x_{ijk}^{\tt de} + a_l\right) + b_m
  \end{split}\,,
\end{equation}
where $w_m, m=1,...,K^2$, is the weight of the upsampling kernel, $K$ the upsampling kernel size, $h$ the convolution window size, $C$ the number of input channel dimension of encoder and decoder features, and $d$ the number of compressed channel dimension. $\alpha_{kl}^{\tt en}$ and $\{\alpha_{kl}^{\tt de}, a_l\}$ are the parameters of $1\times1$ convolution specific to encoder and decoder features, respectively, and $\{\beta_{ijlm}, b_m\}$ the parameters of $3\times3$ convolution. Following CARAFE, we set $h=3$, $K=5$, and $d=64$.

\subsubsection{Efficient Implementations of Semi-shift Convolution}
Given the formulation above, here we discuss the efficient implementations of semi-shift convolution. According to Eq.~\eqref{eq:1}, by the linearity of convolution, the two standard convolutions on $2C$-channel features are equivalent to applying two distinct $1\times1$ convolutions to $C$-channel encoder and $C$-channel decoder features, respectively, followed by a shared $3\times3$ convolution and summation. Such decomposition allows us to process encoder and decoder features without matching their resolution explicitly. However, we still need to 
address the mismatch implicitly. There are two strategies: i) downsampling the high-res encoder output to match the low-res decoder one, or ii) upsampling the low-res decoder output to match the high-res encoder one.

\begin{figure}[!t]
	\centering
	\includegraphics[width=\linewidth]{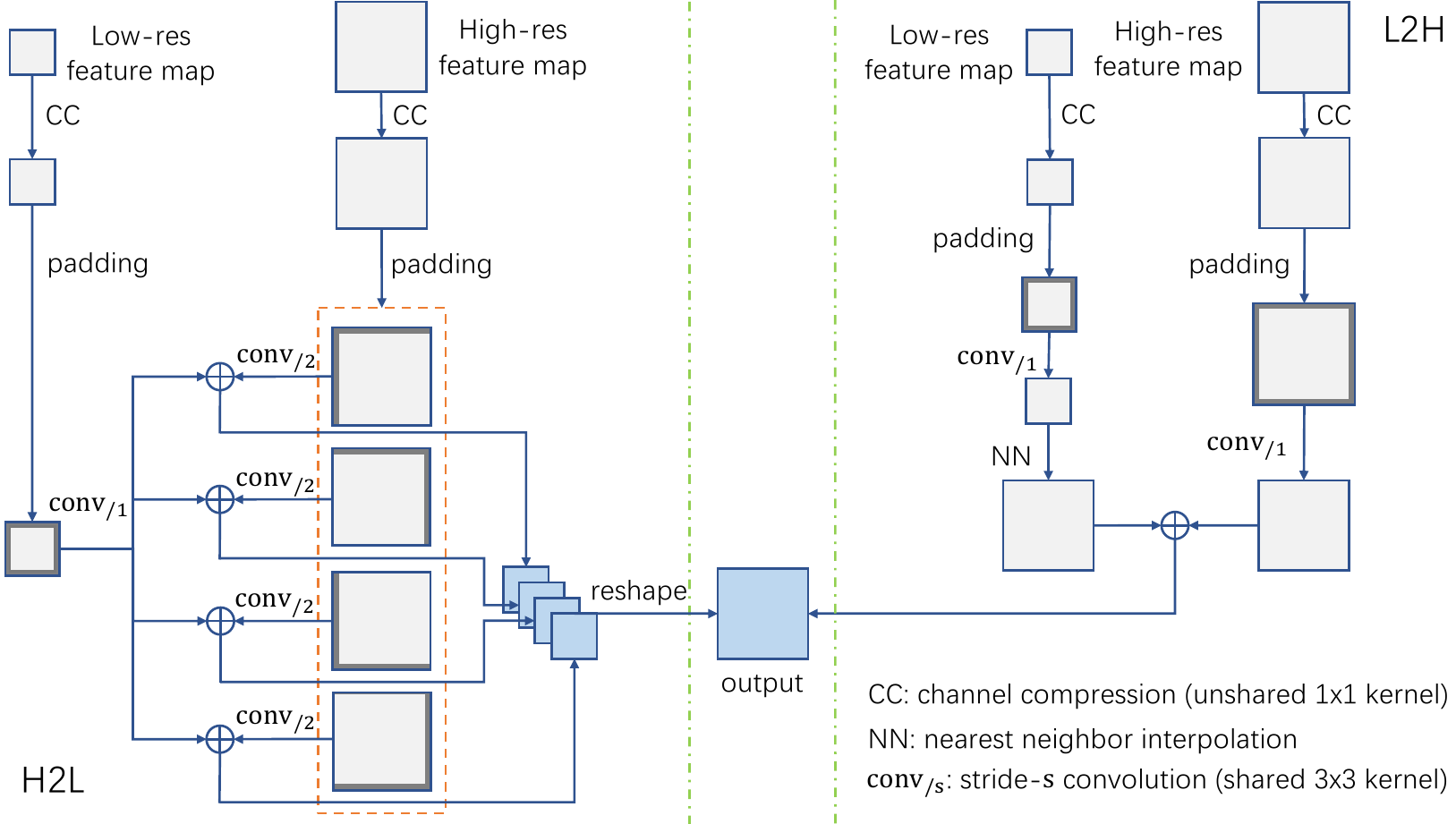}
	\caption{\textbf{Fast implementations of semi-shift convolution.} We present two forms of fast implementations: (left: H2L) high resolution matches low resolution, which is presented in our conference version \citep{lu2022fade}, and (right: L2H) low resolution matches high resolution, which is more memory efficient.}
	\label{fig:fast_implementation}
\end{figure}

To process the whole feature map following the first strategy, the window can move $s$ steps on encoder features but only $\lfloor s/2 \rfloor$ steps on decoder features. This is why the operator is given the name `semi-shift convolution'. We split the process to $4$ sub-processes; each sub-process focuses on the top-left, the top-right, the bottom-left, and the bottom-right window, respectively. Different sub-processes have similar prepossessing strategies. For example, for the top-left sub-process, we add full zero padding to the decoder feature, but only pad the top and left side of the encoder feature. Then all the top-left window correspondences can be satisfied by setting convolutional stride of $1$ for the decoder feature and of $2$ for the encoder feature. Finally, after a few memory operations, the four sub-outputs can be reassembled to the (unnormalized) upsampling kernel. 
This process is illustrated in the left of Fig.~\ref{fig:fast_implementation}, which can be called the high-to-low (H2L) implementation.

The H2L implementation above is provided in our conference version \citep{lu2022fade}. We later notice that the key characteristic of semi-shift convolution lies in \textit{the same decoder feature point corresponds to $4$ encoder feature points}, which shares the same spirit of NN interpolation. Following this interpretation, we provide a more efficient implementation with less use of memory, as shown in the right of Fig.~\ref{fig:fast_implementation}, named the low-to-high (L2H) implementation. First, unshared $1\times 1$ convolutions are used to compress the encoder and decoder features, respectively. Then the shared $3\times3$ convolution is applied, 
of which the decoder feature is NN-interpolated to the size of the encoder one. Finally they are summed 
to obtain the (unnormalized) kernel.

Both implementations can be implemented within the standard ${\tt PyTorch}$ library. 
In the H2L implementation, the kernel $\mathcal{W}_i$ of the $i$-th sub-process (with specific padding applied), $i=1,2,3,4$, takes the form
\begin{equation}\label{eq:3}\small
    \mathcal{W}_i = \tt conv_{/2}(CC(\mathcal{X}_{en}, \theta_{en}), \theta)+\tt conv_{/1}(CC(\mathcal{X}_{de}, \theta_{de}), \theta)\,,
\end{equation}
where $\tt conv_{/s}(\mathcal{X}, \theta)$ denotes the stride-$s$ $3\times3$ convolution over the feature map $\mathcal{X}$, parameterized by $\theta$. $\tt CC$ is the channel compressor implemented by $1\times1$ convolution. $\tt \mathcal{X}_{en}$ and $\tt \mathcal{X}_{de}$ are the encoder and the decoder feature, respectively. Note that, the parameters $\tt \theta_{en}$ and $\tt \theta_{de}$ in $\tt CC$ are different, while the parameters in $\tt conv_{/1}$ and $\tt conv_{/2}$ are the same $\theta$. The four $\mathcal{W}_i$'s need to be aggregated and reshaped to form the full kernel $\mathcal{W}$.

In contrast, the L2H implementation does not require sub-process division and computes the full kernel $\mathcal{W}$ directly. It can be formulated as
\begin{equation}\label{eq:4}\small
    \tt \mathcal{W} = \tt conv_{/1}(CC(\mathcal{X}_{en}, \theta_{en}), \theta)+ NN (\tt conv_{/1}(CC(\mathcal{X}_{de}, \theta_{de}), \theta))\,,
\end{equation}
where $\tt NN$ is the $\times2$ NN interpolation operator.

\paragraph{SemiShift-Lite and FADE-Lite.} We also investigate a simplified 
variant of semi-shift convolution, which uses depthwise convolution to further reduce the computational complexity, named 
SemiShift-Lite. Specifically, SemiShift-Lite sets $d=K^2$ and adopts $3\times3$ depthwise convolution to encode the local information. Its whole number of parameters is $2CK^2+9K^2$. The use of SemiShift-Lite also leads to a lightweight variant of FADE, \textit{i.e.}, FADE-Lite. We use this variant to show that the task-agnostic property indeed comes with the careful treatment of encoder and decoder features, even with much less parameters. When $C=256$, $d=64$, and $K=5$, despite FADE-Lite only includes $27.6\%$ parameters of its standard version FADE, we observe that FADE-Lite is still task-agnostic and outperforms most upsampling operators (see Section~\ref{sec:application} for details).


\subsection{Extracting Fine Details from Encoder Features}
\label{subsec:gate}

Here we further introduce a gating mechanism to complement fine details from encoder features to upsampled features. We again use some experimental observations to motivate our design. We use a binary image segmentation dataset, Weizmann Horse \citep{borenstein2002class}. The reasons for choosing this dataset are two-fold: 
(1) the visualization is made simple; 
(2) the task is simple such that the impact of feature quality can be neglected. 
When all baselines have nearly perfect region predictions, the difference in detail prediction can be amplified. We use SegNet pretrained on ImageNet as the baseline and alter only the upsampling operators.
Results are listed in Table \ref{tab:horse}. 
An interesting phenomenon is that CARAFE works almost the same as NN interpolation and even falls behind the default unpooling and IndexNet. An explanation is that the dataset is too simple such that the region smoothing property of CARAFE is wasted, but recovering details matters. 

\begin{table}[!t]\small
    \caption{Results on the Weizmann Horse dataset}
    \label{tab:horse}
    \centering
    \renewcommand{\arraystretch}{1.2} 
    \addtolength{\tabcolsep}{42pt}
    \begin{tabular}{@{}lc@{}}
        \toprule
        SegNet -- baseline & mIoU\\
        \midrule
        Unpooling & 93.42\\
        IndexNet \citep{lu2019indices} & 93.00\\
        NN & 89.15\\ 
        CARAFE \citep{jiaqi2019carafe} & 89.29\\
        NN + Gate & 95.26\\
        CARAFE + Gate & 95.25\\
        \bottomrule
    \end{tabular}
\end{table}

\begin{figure}[!t]
    \centering
    \includegraphics[width=1\linewidth]{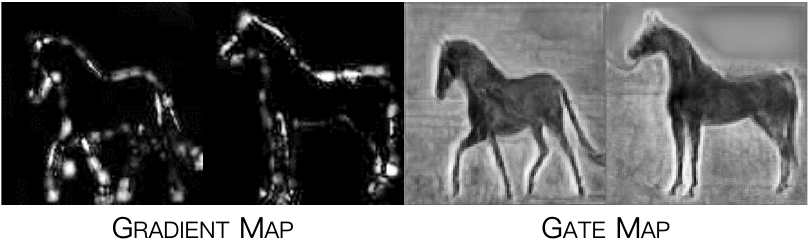}
    \caption{Gradient maps and gate maps of horses.}
    \label{fig:horse}
\end{figure}

A common sense in segmentation is that, the interior of a certain class would be learned fast, while mask boundaries are difficult to predict.  
This can be observed from the gradient maps w.r.t.\ an intermediate decoder layer, as shown in Fig.~\ref{fig:horse}. During the middle stage of training, most responses are near boundaries. 
Now that gradients reveal the demand of detail information, feature maps would also manifest this requisite with some distributions, e.g., in multi-class semantic segmentation a confident class prediction in a region would be a unimodal distribution along the channel dimension, and an uncertain prediction around boundaries would likely be a bimodal distribution. Hence, we assume that all decoder layers have gradient-imposed distribution priors and can be encoded to inform the requisite of detail or semantic information. In this way fine details can be chosen from encoder features without hurting the semantic property of decoder features. Hence, instead of directly skipping encoder features as in feature pyramid networks (FPNs) \citep{lin2017feature}, we introduce a naive gating mechanism following existing ideas \citep{cho2014properties,li2020gated,li2023sfnet} to refine upsampled features using encoder features, conditioned on decoder features. The gate is generated through a $1\times1$ convolution layer, a NN interpolation layer, and a $\tt sigmoid$ function. As shown in Fig.~\ref{fig:pipeline}(c), the decoder feature first goes through the gate generator, and the generator then outputs a gate map instantiated in Fig.~\ref{fig:horse}. Finally, the gate map $G$ modulates the encoder feature $\mathcal F_{\tt encoder}$ and the upsampled feature $\mathcal F_{\tt upsampled}$ to generate the final refined feature $\mathcal F_{\tt refined}$ as
\begin{equation} \small
    \label{eq:5}
  \mathcal F_{\tt refined} = \mathcal F_{\tt encoder} \cdot G + \mathcal F_{\tt upsampled} \cdot (1-G)\,.
\end{equation}
From Table~\ref{tab:horse}, the gate works on both NN and CARAFE.

We remark that our initial motivation for developing the gating mechanism comes from semantic segmentation and image matting tasks. In semantic segmentation, the model outputs a set of logits and uses {\rm argmax} to select one channel as the predicted class. This form of prediction renders the model working in a one-class-one-value manner. To preserve this manner, we expect the gate to extract only the details that require from the encoder (Fig.~\ref{fig:horse}) and to influence the decoder feature as less as possible. Similarly in matting, despite the number of classes can be considered to be infinity, the model still follows the one-class-one-value paradigm. However, in instance-sensitive tasks, such as object detection, given the one-class-one-value feature maps, one cannot tell the instance difference with {\rm argmax}. In addition, object detection is rather different from semantic segmentation, where high-res features are responsible for precise localization, so in \citep{lin2017feature} the FPN is adopted to improve Faster-RCNN \citep{ren2015faster}. For the reasons above, gating, as a mechanism strengthening decoder features, may not tackle the improvement for localization. In this case, FADE without gating, denoted by FADE (G=1), would be a better choice. We will discuss more in the experiments on object detection (Section~\ref{subsec:object_detection}) and instance segmentation (Section~\ref{subsec:instance_segmentation}).

\begin{table*}[!t] \footnotesize
    \caption{Semantic segmentation and image matting results on the ADE20K and Adobe Composition-1K data sets. For IndexNet, we use the `M2O' version in matting and `HIN' in segmentation following \citep{lu2022index}. GFLOPs and Param indicate the additional floating-point calculations and additional number of parameters compared with the bilinear baseline. Best performance is in \textbf{boldface} and second best is \underline{underlined}}
    \centering
    \renewcommand{\arraystretch}{1.2}
    \addtolength{\tabcolsep}{-1.5pt}
    \begin{tabular}{@{}l cc cl|cccc cl@{}}
    \toprule
        SegFormer-B1/ & \multicolumn{4}{c|}{$\tt Segmentation$ -- $\tt accuracy$ $\tt metric\uparrow$} & \multicolumn{6}{c}{$\tt Matting$ -- $\tt error$ $\tt metric\downarrow$}  \\
        A2U Matting-R34 & mIoU & bIoU & GFLOPs & Params & SAD & MSE & Grad & Conn & GFLOPs & Params\\
        \midrule
        Bilinear & 41.68 & 27.80 & 15.9 & 13.7M & 37.31 & 0.0103 & 21.38 & 35.39 & 8.6 & 8.1M\\
        CARAFE \citep{jiaqi2019carafe} & 42.82 & 29.84 & +1.5 & +0.4M & 41.01 & 0.0118 & 21.39 & 39.01 & +6.0 & +0.3M\\
        IndexNet \citep{lu2019indices} & 41.50 & 28.27 & +30.7 & +12.6M & 34.28 & 0.0081 & 15.94 & 31.91 & +31.7 & +12.3M\\
        A2U \citep{dai2021learning} & 41.45 & 27.31 & +0.4 & +0.1M & 32.15 & 0.0082 & 16.39 & 29.25 & +0.7 & +38K\\
        SAPA \citep{lu2022sapa} & 43.20 & 30.96 & +1.7 & +0.2M & \underline{31.19} & 0.0079 & 15.48 & 28.30 & +1.8 & +0.1M \\
        FADE & \textbf{44.41} & \textbf{32.65} & +2.7 & +0.3M & \textbf{31.10} & \textbf{0.0073} & \textbf{14.52} & \textbf{28.11} & +8.9 & +0.1M\\
        FADE-Lite & \underline{43.49} & \underline{31.55} & +0.9 & +80K & 31.36 & \underline{0.0075} & \underline{14.83} & \underline{28.21} & +1.5 & +27K\\
    \bottomrule
    \end{tabular}
    \label{tab:performance}
\end{table*}

\begin{table}[!t]
\caption{Semantic segmentation results on the ADE20K data set with different SegFormer backbones}
\label{tab:segformer_scaling}
\centering
\renewcommand{\arraystretch}{1.2}
\addtolength{\tabcolsep}{-3.5pt}
\begin{tabular}{lclll}
\toprule
SegFormer & backbone & params & mIoU & bIoU\\
\midrule 
Bilinear & B1 & 13.7M & 41.68 & 27.80 \\
FADE & B1 & +0.4M & 44.41 (+2.73) & 32.65 (+4.85) \\
\rowcolor{gray!10}
Bilinear & B3 & 47.3M & 49.04 & 35.24 \\
\rowcolor{gray!10}
FADE & B3 & +0.7M & 49.05 (+0.01) & 36.79 (+1.55) \\
Bilinear & B4 & 64.1M & 49.93 & 35.63 \\
FADE & B4 & +0.7M & 50.11 (+0.18) & 37.39 (+1.76) \\
\rowcolor{gray!10}
Bilinear & B5 & 84.7M & 51.00 & 37.81 \\
\rowcolor{gray!10}
FADE & B5 & +0.7M & 50.90 (-0.10) & 38.91 (+1.10) \\
\bottomrule
\end{tabular}
\label{tab:segformer_backbone}
\end{table}

\section{Applications}
\label{sec:application}

Here we demonstrate the applications and the task-agnostic property of FADE on various 
dense prediction tasks, including semantic segmentation, image matting, object detection, instance segmentation, and monocular depth estimation.
In particular, we focus our experiments on segmentation to analyze the the upsampling behaviors of FADE from different aspects and design ablation studies to justify our design choice on FADE. 

\subsection{Semantic Segmentation}
Semantic segmentation is region-sensitive. 
To prove that FADE is architecture-independent, SegFormer \citep{xie2021segformer} and UPerNet \citep{xiao2018unified} 
are chosen as transformer and convolutional baselines, respectively.

\subsubsection{Data Set, Metrics, Baseline, and Protocols}
We use the ADE20K dataset \citep{zhou2017scene}. 
It covers $150$ fine-grained semantic concepts, including $20,210$ training images and $2,000$ validation images. In addition to reporting the standard mask IoU (mIoU) \citep{everingham2010pascal}, we also report the boundary IoU (bIoU) \citep{cheng2021boundary} to assess the boundary quality.

SegFormer-B1 \citep{xie2021segformer} is first evaluated. 
We keep the default model architecture in SegFormer except for modifying the upsampling stages in the MLP head. In particular, feature maps of each scale need to be upsampled to $1/4$ of the original image. Therefore, there are $3+2+1=6$ upsampling stages in all. All training settings and implementation details are kept the same as in \citep{xie2021segformer}. Since SegFormer follows a `fuse-and-concatenate' manner, where the feature maps are all upsampled to the max-resolution one, we verify two styles of upsampling strategies: direct upsampling and $2$ by $2$ iterative upsampling. We also test B3, B4, and B5 versions of SegFormer to see if a similar boost could be observed on stronger backbones. In addition, considering that stronger backbones often produce better feature quality, this also allows to see whether feature upsampling still contributes to improved feature quality on stronger backbones.

For UPerNet \citep{xiao2018unified}, 
we use the implementation provided by \texttt{mmsegmentation}.\footnote{https://github.com/open-mmlab/mmsegmentation} We use the ResNet-50 and ResNet-101 backbones and modify the upsampling operators in the FPN and train the model with $80$ K iterations. The original skip connection is removed 
due to the inclusion of the gating mechanism. Because FADE upsamples by $\times2$ times of the input at once, we use the aligned resizing in inference to match the resolution. Other settings are kept the same.  

\subsubsection{Semantic Segmentation Results}
\begin{figure*}[!t]
	\centering
	\includegraphics[width=\linewidth]{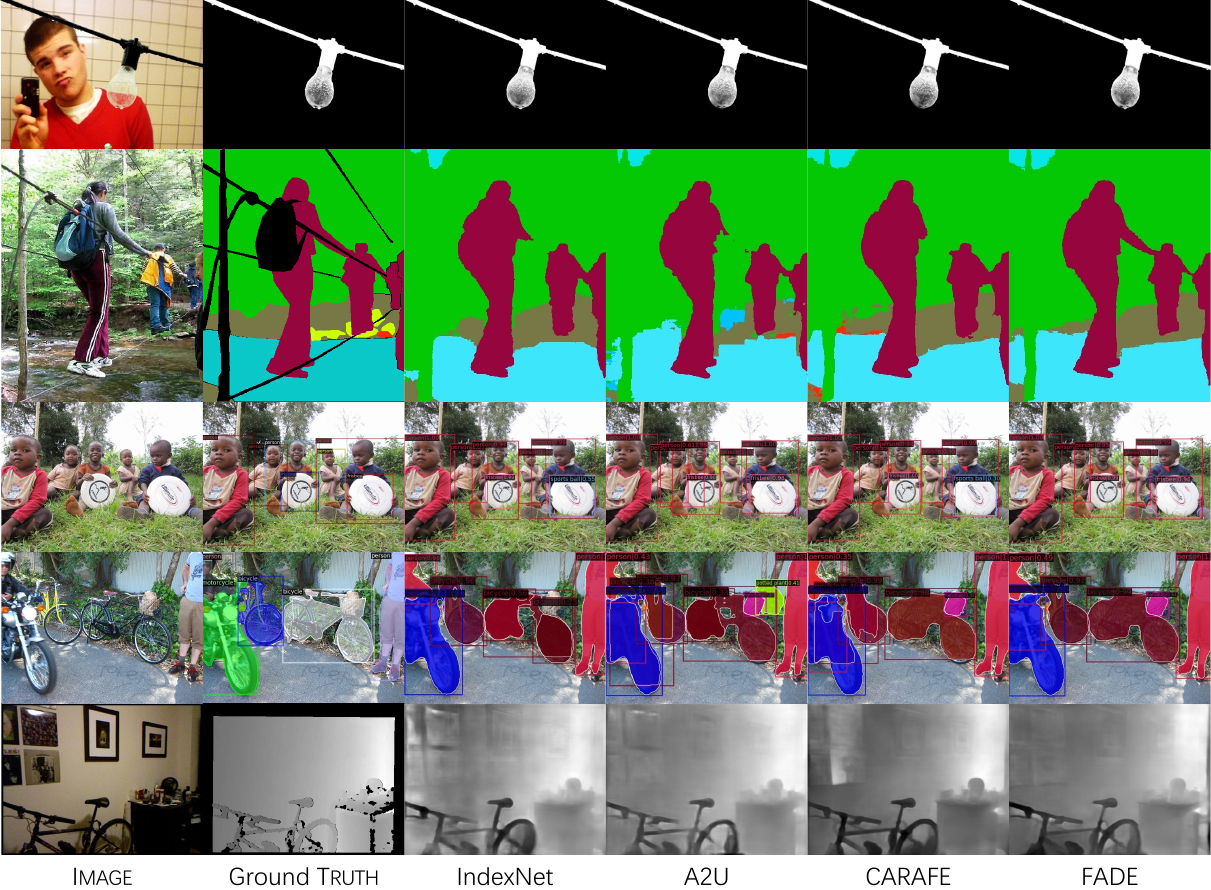}
	\caption{\textbf{Qualitative results of different upsampling operators on different dense prediction tasks.} Among all competitors, only FADE produces visually pleasing visualizations on both region- and detail-sensitive tasks, \textit{e.g.}, the water drops under the bulb, the hand on the rope, and the (generally) smooth depth values of the wall.}
	\label{fig:fade_visual}
\end{figure*}

Quantitative results of different upsampling operators are reported in Table~\ref{tab:performance}. 
FADE is the best performing operator on both mIoU and bIoU metrics. In particular, it improves over the Bilinear baseline by a large margin, with $+2.73$ mIoU and $+4.85$ bIoU. Qualitative results are shown in Figures~\ref{fig:seg_matte_intro} and~\ref{fig:fade_visual}. FADE generates high-quality predictions both within mask regions and near mask boundaries.

\textit{Stronger Backbones.} We also test stronger backbones on SegFormer, including the B3, B4, and B5 versions. 
From Table~\ref{tab:segformer_backbone}, when stronger backbones are used, we observe both mIoU and bIoU improve (B1$\rightarrow$B3, B3$\rightarrow$B4, and B4$\rightarrow$B5). However, on B3, B4, and B5, the benefits of FADE are almost invisible in terms of mIoU, which suggests improved feature quality brought by improved backbones have addressed many misclassifications that upsampling can amend, particularly for interior regions. Yet, steady boosts in bIoU ($>1$) can still be observed. This means improved features only address the boundary errors to a certain degree (cf. bIoU improvements in B1$\rightarrow$B3 vs.\ that in B3$\rightarrow$B4), and FADE can still improve feature quality near mask boundaries. Our evaluations connote improved feature upsampling indeed makes a difference, particularly being useful for resource-constrained applications where a model has limited capacity.

\textit{Upsampling Styles.} We also explore two styles of upsampling in SegFormer: direct upsampling and iterative $\times2$ upsampling. From Table~\ref{tab:segformer_style} we can see that iterative upsampling is better than the direct one in performance. Compared with CARAFE, FADE is more sensitive to the upsampling style, which implies the occurrence of features of different scales matters.

\textit{Applicability to CNN Architecture.} We further evaluate FADE on UPerNet. Results are shown in Table~\ref{tab:UPerNet}. Compared with Bilinear, FADE boosts around $+1\%$ mIoU and outperforms the strong baseline CARAFE with ResNet-50, which confirms the efficacy of FADE for the FPN architecture. On the ResNet-101 backbone, FADE also works, and we observe a even more significant improvement in bIoU, which suggests FADE is good at amending boundary errors.

\begin{figure}[!t]
	\centering
	\includegraphics[width=\linewidth]{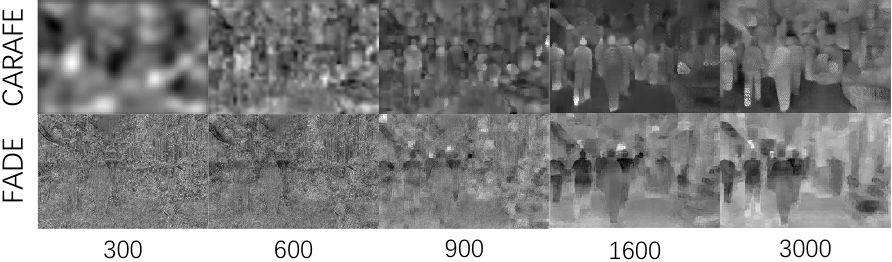}
	\caption{\textbf{Learned upsampled feature maps with increased iterations.} The learning process between CARAFE and FADE is different. FADE first delineates the outlines of objects and then fills the interior regions, while CARAFE starts from the interior and then spreads outside.}
	\label{fig:learning_process}
\end{figure}

\textit{Visualization of Learned Upsampling.} We also visualize the learning process of CARAFE and FADE with increased iterations. From Fig.~\ref{fig:learning_process}, we can see that the two upsampling operators have different behaviors: FADE first learns to delineate the outlines of objects and then fills the interior regions, while CARAFE focuses on the interior initially and then spreads outside slowly. We think the reason is that the gating mechanism is relatively simple and learns fast. By the way, one can see that there are checkerboard artifacts in the visualizations of CARAFE (on the leg of the bottom left person) due to the adoption of Pixel Shuffle. Such visualizations suggest that upsampling can significantly affect the quality of features. While there is no principal rule on what could be called `good features', feature visualizations still proffer a good basis of the feature quality, and one at least can sense where is wrong when clear artifacts present in visualizations.

\subsection{Image Matting}
Our second task is image matting \citep{xu2017deep}. Image matting is a typical detail-sensitive task. It requires a model to estimate an accurate alpha matte that smoothly splits foreground from background. Since ground-truth alpha mattes can exhibit significant differences among local regions, estimations are sensitive to a specific upsampling operator used \citep{lu2019indices,dai2021learning}.

\begin{table}[!t]
\caption{SegFormer with direct or iterative upsampling of FADE and CARAFE. 
}
\centering
\renewcommand{\arraystretch}{1.2}
\addtolength{\tabcolsep}{10pt}
\begin{tabular}{@{}lcc@{}}
\toprule
SegFormer-B1 & \multicolumn{2}{c}{mIoU} \\
 & direct & iterative \\
\midrule
CARAFE \citep{jiaqi2019carafe} & 42.67 & 42.82\\
FADE   & 42.89 & 44.41\\
\bottomrule
\end{tabular}
\label{tab:segformer_style}
\end{table}

\begin{table}[!t]
\caption{Semantic segmentation results with UPerNet. Best performance is in \textbf{boldface}}
\centering
\renewcommand{\arraystretch}{1.2}
\addtolength{\tabcolsep}{0pt}
\begin{tabular}{@{}lccc@{}}
\toprule
UPerNet & backbone & mIoU & bIoU\\
\midrule
Bilinear & R50 & 41.09 & 28.04\\
CARAFE \citep{jiaqi2019carafe} & R50 & 41.49 & 28.29\\
FADE & R50 & \textbf{42.18} & \textbf{28.72}\\
\midrule
Bilinear & R101 & 43.33 & 30.21 \\
FADE & R101 & \textbf{44.27} & \textbf{31.31} \\
\bottomrule
\end{tabular}
\label{tab:UPerNet}
\end{table}

\subsubsection{Data Set, Metrics, Baseline, and Protocols} 
We conduct experiments on the Adobe Image Matting dataset \citep{xu2017deep}, whose training set has $431$ unique foreground objects and ground-truth alpha mattes. Following \citep{dai2021learning}, instead of compositing each foreground with fixed $100$ background images chosen from MS COCO \citep{lin2014microsoft}, we randomly choose background images in each iteration and generate composited images on-the-fly. The Composition-1K testing set has $50$ unique foreground objects, and each is composited with $20$ background images from PASCAL VOC \citep{everingham2010pascal}. We report the widely used Sum of Absolute Differences (SAD), Mean Squared Error (MSE), Gradient (Grad), and Connectivity (Conn) \citep{rhemann2009perceptually}. 

A2U Matting \citep{dai2021learning} is adopted as the baseline. Following \citep{dai2021learning}, the baseline network adopts a backbone of the first $11$ layers of ResNet-34 with in-place activated batchnorm \citep{bulo2018place} and a decoder consisting of a few upsampling stages with shortcut connections. Readers can refer to \citep{dai2021learning} for the detailed architecture. We use max pooling in downsampling stages when applying FADE as the upsampling operator to train the model, and cite the results of other upsampling operators from A2U Matting \citep{dai2021learning}. We strictly follow the training configurations and data augmentation strategies used in \citep{dai2021learning}.

\begin{table*}[!t]
\caption{Object detection results with Faster R-CNN on MS-COCO. The `HIN' version of IndexNet is used. Best performance is in \textbf{boldface} and second best is \underline{underlined}}
\centering
\renewcommand{\arraystretch}{1.2}
\addtolength{\tabcolsep}{5.5pt}
\begin{tabular}{@{}lclcccccc@{}}
\toprule
Faster RCNN \citep{ren2015faster} & backbone & Params & $AP$    & $AP_{50}$ & $AP_{75}$ & $AP_S$  & $AP_M$  & $AP_{L}$  \\
\midrule
FA2M \citep{wu2022fsanet} & R50 & +23K & 37.9 & 58.8 & 40.9 & 22.1 & 41.7 & 48.8 \\
FAM \citep{li2020semantic} & R50 & +0.8M & 37.8 & 58.6 & 41.0 & 21.8 & 41.2 & 48.8 \\
GD-FAM \citep{li2023sfnet} & R50 & +0.5M & 38.1 & 59.2 & 41.3 & 22.7 & 41.5 & \underline{49.6} \\
\midrule 
Nearest & R50 & 46.8M & 37.4 & 58.1 & 40.4 & 21.2 & 41.0 & 48.1 \\
CARAFE \citep{jiaqi2019carafe} & R50 & +0.3M & \textbf{38.6} & \textbf{59.9} & \textbf{42.2} & \textbf{23.3} & \textbf{42.2} & \textbf{49.7} \\
IndexNet \citep{lu2022index} & R50 & +8.4M & 37.6 & 58.4 & 40.9 & 21.5 & 41.3 & 49.2 \\
A2U \citep{dai2021learning} & R50 & +0.1M & 37.3 & 58.7 & 40.0 & 21.7 & 41.1 & 48.5 \\
SAPA \citep{lu2022sapa} & R50 & +0.1M & 37.8 & 59.2 & 40.6 & 22.4 & 41.4 & 49.1 \\
FADE & R50 & +0.2M & 37.8 & 58.8 & 40.8 & 21.2 & 41.2 & 49.4 \\
FADE (G=1) & R50 & +0.2M & \underline{38.5} & \underline{59.6} & \underline{41.8} & \underline{23.1} & \textbf{42.2} & 49.3 \\
FADE-Lite (G=1) & R50 & +52K & 38.3 & 59.5 & 41.7 & 22.4 & \underline{41.9} & \textbf{49.7} \\
\midrule 

Nearest & R101 & 65.8M & 39.4 & 60.1 & 43.1 & 22.4 & 43.7 & 51.1 \\
FADE (G=1) & R101 & +0.2M & 40.0 & 61.0 & 43.3 & 23.6 & 44.0 & 51.1 \\

\bottomrule
\end{tabular}
\label{tab:object_detection}
\end{table*}

\subsubsection{Image Matting Results} We compare FADE with other state-of-the-art upsampling operators. Quantitative results are also shown in Table~\ref{tab:performance}. Akin to segmentation, FADE consistently outperforms other competitors in all metrics, with also few additional parameters. 
Note that IndexNet and A2U are strong baselines that are delicately designed upsampling operators for image matting. Also the worst performance of CARAFE indicates that upsampling with only decoder features is not sufficient to recover details. Compared with standard bilinear upsampling, FADE invites $16\%\sim32\%$ relative improvements, which suggests a simple upsampling operator can make a difference. Our community may shift more attention to upsampling. 
Additionally, it is worth noting that FADE-Lite also outperforms other prior operators, and particularly, surpasses the strong baseline A2U with even less parameters. Qualitative results are shown in Figures~\ref{fig:seg_matte_intro} and~\ref{fig:fade_visual}. FADE generates high-fidelity alpha mattes.

\textit{Task-Agnostic Property.} By comparing different upsampling operators across both segmentation and matting, FADE is the only operator that exhibits the task-agnostic property. A2U is the previous best operator in matting, but turns out to be the worst one in segmentation. CARAFE is the previous best operator in segmentation, but the worst one in matting. This implies that current dynamic operators still have certain weaknesses to achieve task-agnostic upsampling. In addition, FADE-Lite also exhibits the task-agnostic property (being the consistent second best in both tasks in all metrics), which suggests such a property is insensitive to the number of parameters.

\subsection{Object Detection}
\label{subsec:object_detection}

The third task is object detection \citep{ren2015faster}. Object detection addresses where and what objects are with category-specific bounding boxes. It is 
a mainstream dense prediction problem. Addressing `what' is a recognition problem, while addressing `where' requires precise localization in feature pyramids. Upsampling is therefore essential to acquire high-res feature maps. 

\subsubsection{Data Set, Metrics, Baseline, and Protocols}
We use the MS COCO dataset \citep{lin2014microsoft} and report the standard $AP$, $AP_{50}$, $AP_{75}$, $AP_S$, $AP_M$, and $AP_L$. We use Faster R-CNN as the baseline and replace the default NN interpolation with other upsampling operators. We follow the Faster R-CNN implementation provided by \texttt{mmdetection}\footnote{https://github.com/open-mmlab/mmdetection} and only modify the upsampling stages in FPN. Note that, the original skip connection in FPN is removed due to the inclusion of the gating mechanism. All other settings remain unchanged. We evaluate on both ResNet-50 and ResNet-101 backbones. Moreover, since the FPN is used, in addition to the dynamic upsampling operators, we also compare with some feature alignment modules designed for FPN, including the FA$^2$M used in FSANet \citep{wu2022fsanet}, the FAM used in SFNet \citep{li2020semantic}, and the GD-FAM used in SFNet-Lite \citep{li2023sfnet}.

\begin{figure}[!t]
	\centering
	\includegraphics[width=\linewidth]{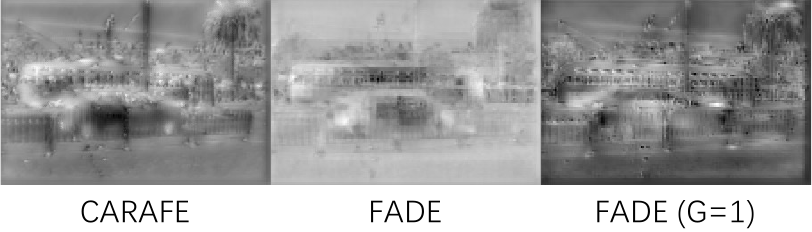}
	\caption{Upsampled feature maps of different upsampling operators on Faster R-CNN.}
	\label{fig:faster-rcnn_feat_map}
\end{figure}

\subsubsection{Object Detection Results}
Quantitative and qualitative results are shown in Table~\ref{tab:object_detection} and Fig.~\ref{fig:fade_visual}, respectively. We find that, while FADE still improves detection performance, it is not at a level comparable to CARAFE. However, when setting the gate $G=1$ in FADE, the performance improves from $37.8$ $AP$ to $38.5$ $AP$, approaching to CARAFE. We are interested to know why. After a careful check at the upsampled feature map (Fig.~\ref{fig:faster-rcnn_feat_map}), we see that the detector favors more detailed upsampled features than blurry ones (CARAFE vs.\ FADE). Perhaps details in features can benefit precise localization of bounding boxes. In the use of CARAFE, high-res encoder features are directly skipped in the FPN. In contrast, FADE uses a gate to control of pass the encoder features. The resulting features of FADE show that the gate does not work as expected: the decoder features dominate in the output. Why does not the gate work? We believe this can boil down to how the detector is supervised. Since the gate predictor has few parameters, the generated gate is mostly affected by the feature map. In semantic segmentation and image matting where per-pixel ground truths are provided, the features can be updated delicately. Yet, in detection where the ground truth bounding boxes are sparse, the feature learning could be coarse, therefore affecting the prediction of the gate. Fortunately, the gating mechanism works in FADE as a post-processing step and can be disabled when unnecessary. In addition, we observe FADE (G=1) outperforms feature alignment modules, which suggests manipulating kernels seems more effective than manipulating features. A plausible explanation is that, feature alignment needs to correct additional artifacts introduced by naive feature upsampling (NN or bilinear upsampling is typically executed before feature alignment is performed).
Moreover, with a stronger backbone ResNet-101, FADE can also boost the performance. This implies that, while a better backbone is often favored, there are still feature issues that cannot be addressed with increased model capacity. In this case, some improved components within the architecture such as improved upsampling may help.


\subsection{Instance Segmentation}
\label{subsec:instance_segmentation}

The forth task is instance segmentation \citep{he2017mask}. Instance segmentation is an extended task of semantic segmentation. In addition to labelling object/scene categories, it needs to further discriminate instances of the same category. It can also be considered a region-sensitive task.

\subsubsection{Data Set, Metrics, Baseline, and Protocols}
Akin to object detection, we use the MS COCO dataset \citep{lin2014microsoft} for instance segmentation and report box $AP$, mask $AP$, and boundary $AP$. Following \citep{jiaqi2019carafe}, we select Mask R-CNN as our baseline and only replace the default NN interpolation with other upsampling operators in the FPN. Since the gate in FADE would reduce to the skip connection when $G=1$ according to Eq.~\eqref{eq:5}, the original skip connection in FPN is removed. We also follow the Mask R-CNN implementation provided by \texttt{mmdetection} and the training setting used in \citep{jiaqi2019carafe}. We test on both ResNet-50 and ResNet-101 backbones. In addition, we also compare against the feature alignment modules as in detection, because Mask R-CNN uses the FPN as well.

\begin{figure}[!t]
	\centering
	\includegraphics[width=\linewidth]{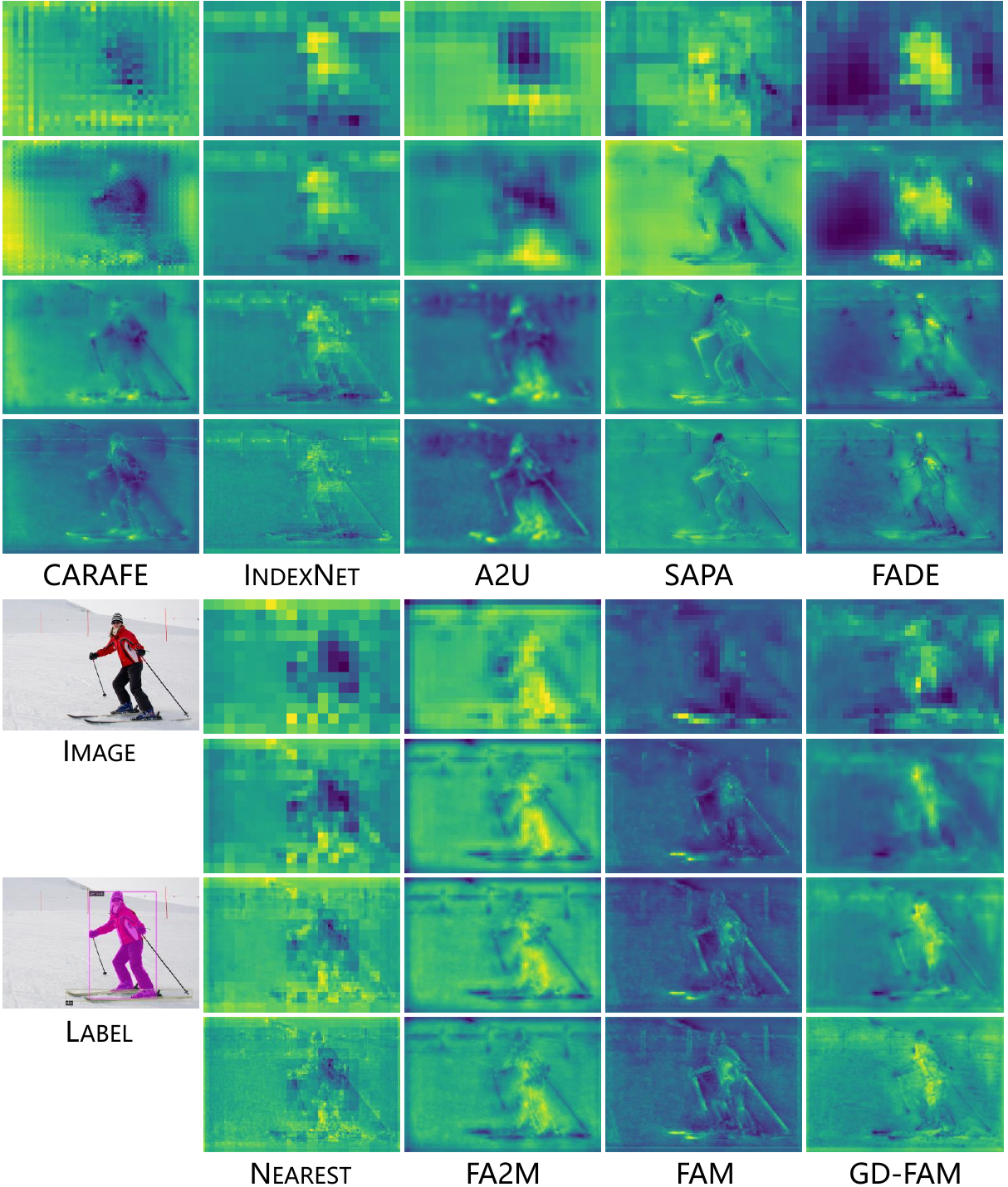}
	\caption{\textbf{Visualizations of upsampled feature maps generated by different methods.} The feature maps are extracted from the output of upsamplers in Mask R-CNN-R50 \citep{he2017mask}. The quality of feature maps generally provides an implication of performance.}
	\label{fig:feature_map}
\end{figure} 

\begin{table*}[!t] \small
\caption{Instance segmentation results with Mask R-CNN (ResNet50 as the backbone) on MS-COCO. Upsampling operators are replaced in FPN. `HIN' version of IndexNet is used. The parameter increment is the same as in Faster R-CNN. Best performance is in \textbf{boldface} and second best is \underline{underlined}}
\centering
\renewcommand{\arraystretch}{1.2}
\addtolength{\tabcolsep}{9.5pt}
\begin{tabular}{@{}lccccccc@{}}
\toprule
Mask R-CNN \citep{he2017mask} & & \multicolumn{6}{c}{Bbox metric}\\
\cmidrule(r){3-8}
Method & backbone & $AP$ & $AP_{50}$ & $AP_{75}$ & $AP_S$  & $AP_M$  & $AP_{L}$  \\
\midrule 
FA2M \citep{wu2022fsanet} & R50 & 38.8 & 59.7 & 42.1 & 22.7 & 42.2 & 50.3 \\
FAM \citep{li2020semantic} & R50 & 38.4 & 59.2 & 41.7 & 21.9 & 41.6 & 49.9 \\
GD-FAM \citep{li2023sfnet} & R50 & 38.7 & 59.4 & 42.2 & 22.4 & 41.7 & 50.7 \\
\midrule
Nearest  & R50 & 38.3 & 58.7 & 42.0 & 21.9 & 41.8 & 50.2 \\
CARAFE \citep{jiaqi2019carafe}   & R50 & \textbf{39.2} & \textbf{60.3} & \textbf{42.9} & \textbf{23.4} & \underline{42.5} & \textbf{51.2} \\
IndexNet \citep{lu2022index} & R50 & 38.4 & 59.1 & 42.1 & 22.1 & 42.0 & 50.2 \\
A2U \citep{dai2021learning}      & R50 & 37.9 & 59.0 & 40.8 & 22.0 & 41.5 & 49.4 \\
SAPA \citep{lu2022sapa} & R50 & 38.7 & 59.7 & 42.2 & 23.1 & 41.8 & 49.9\\
FADE     & R50 & 38.7 & 59.3 & 42.0 & 22.6 & 42.4 & 50.4 \\
FADE (G=1) & R50 & \textbf{39.2} & \textbf{60.3} & \underline{42.7} & \underline{23.2} & \textbf{42.6} & \underline{51.0} \\
FADE-Lite (G=1) & R50 & \underline{38.9} & \underline{60.0} & 42.3 & \underline{23.2} & 42.2 & 50.9 \\
\midrule
Nearest  & R101 & 40.0 & 60.4 & 43.7 & 22.8 & 43.7 & 52.0 \\
FADE (G=1) & R101 & 40.6 & 61.5 & 44.3 & 24.1 & 44.4 & 53.1 \\
\midrule
 & & \multicolumn{6}{c}{Segm metric}\\
\cmidrule(r){3-8}
Method & backbone & $AP$ & $AP_{50}$ & $AP_{75}$ & $AP_S$  & $AP_M$  & $AP_{L}$  \\
\midrule
FA2M \citep{wu2022fsanet} & R50 & \underline{35.3} & 56.4 & \underline{37.6} & 16.8 & 37.6 & \textbf{52.1} \\
FAM \citep{li2020semantic} & R50 & 34.8 & 56.0 & 36.9 & 15.9 & 37.1 & 50.9 \\
GD-FAM \citep{li2023sfnet} & R50 & 35.0 & 56.1 & 37.3 & 16.6 & 37.1 & 51.4 \\
\midrule
Nearest  & R50 & 34.7 & 55.8 & 37.2 & 16.1 & 37.3 & 50.8 \\
CARAFE \citep{jiaqi2019carafe}   & R50 & \textbf{35.5} & \textbf{57.1} & \textbf{37.8} & \textbf{17.3} & \textbf{37.9} & \underline{51.9} \\
IndexNet \citep{lu2022index} & R50 & 34.8 & 55.9 & 37.0 & 16.3 & 37.3 & 51.3 \\
A2U \citep{dai2021learning}      & R50 & 34.4 & 55.7 & 36.8 & 15.7 & 37.1 & 50.6 \\
SAPA \citep{lu2022sapa} & R50 & \underline{35.3} & 56.7 & \underline{37.6} & 16.9 & \textbf{37.9} & 50.7\\
FADE     & R50 & 34.7 & 55.9 & 36.9 & 15.9 & 37.2 & 51.0 \\
FADE (G=1) & R50 & 35.2 & \underline{56.8} & 37.5 & 16.8 & \underline{37.7} & 51.4 \\
FADE-Lite (G=1) & R50 & \underline{35.3} & 56.7 & \underline{37.6} & \underline{17.2} & \underline{37.7} & 51.7\\
\midrule
Nearest & R101 & 36.0 & 57.6 & 38.5 & 16.5 & 39.3 & 52.2 \\
FADE (G=1) & R101 & 36.4 & 58.0 & 38.9 & 17.4 & 39.3 & 53.3 \\
\midrule
 & & \multicolumn{6}{c}{Boundary metric}\\
\cmidrule(r){3-8}
Method & backbone & $AP$ & $AP_{50}$ & $AP_{75}$ & $AP_S$  & $AP_M$  & $AP_{L}$  \\
\midrule
FA2M \citep{wu2022fsanet} & R50 & 21.1 & 46.2 & \underline{16.8} & 16.8 & 31.4 & \textbf{20.6} \\
FAM \citep{li2020semantic} & R50 & 20.7 & 45.8 & 16.4 & 15.9 & 31.0 & 19.9 \\
GD-FAM \citep{li2023sfnet} & R50 & 20.9 & 46.2 & 16.3 & 16.5 & 30.9 & 20.1 \\
\midrule
Nearest  & R50 & 20.7 & 45.5 & 16.6 & 16.0 & 31.2 & 20.0 \\
CARAFE \citep{jiaqi2019carafe}   & R50 & \textbf{21.3} & \textbf{47.0} & \textbf{16.9} & \textbf{17.3} & \underline{31.5} & \textbf{20.6} \\
IndexNet \citep{lu2022index} & R50 & 20.8 & 45.8 & 16.4 & 16.3 & 31.2 & 20.3 \\
A2U \citep{dai2021learning}      & R50 & 20.4 & 45.1 & 15.9 & 15.7 & 30.8 & 19.7 \\
SAPA \citep{lu2022sapa} & R50 & 21.1 & 46.4 & 16.6 & 16.8 & \textbf{31.7} & 19.6\\
FADE     & R50 & 20.7 & 45.7 & 16.3 & 15.9 & 31.1 & 20.3 \\
FADE (G=1) & R50 & \underline{21.2} & 46.6 & 16.7 & 16.8 & 31.4 & \underline{20.4} \\
FADE-Lite (G=1) & R50 & \underline{21.2} & \underline{46.7} & 16.6 & \underline{17.1} & \underline{31.5} & \textbf{20.6} \\
\midrule
Nearest & R101 & 22.1 & 48.1 & 17.6 & 16.4 & 33.0 & 21.5 \\
FADE (G=1) & R101 & 22.2 & 48.5 & 17.6 & 17.4 & 33.0 & 21.6 \\
\bottomrule
\end{tabular}
\label{tab:instance_segmentation}
\end{table*}

\subsubsection{Instance Segmentation Results}

Quantitative and qualitative results are shown in Table~\ref{tab:instance_segmentation} and Fig.~\ref{fig:fade_visual}, respectively. We have similar observations to object detection: i) the standard implementation of FADE only shows marginal improvements; ii) FADE without gating works better than FADE and is on par with CARAFE. Compared with other tasks, all upsampling operators have limited improvements ($<1$) in terms of mask AP. A reason may be the limited output resolution ($28\times28$) of the mask head. In this case, the benefits of improved boundary delineation of upsampling may not be revealed, which can also be observed from the marginal improvements on the boundary $AP$. Indeed the more significant relative improvements on box $AP$ than mask $AP$ indicate that the improved mask AP could be mostly due to the improved detection performance. Nevertheless, FADE without gating could still be a preferable choice if taking its task-agnostic property into account. With a stronger backbone ResNet-101, FADE invites an improvement of $0.6$ box $AP$ and $0.4$ mask $AP$, which provides a similar boost as ResNet-50. Compared with feature alignment modules, dynamic upsampling operators generally work better. From the visualizations of feature maps in Fig.~\ref{fig:feature_map}, one can see that, despite being empirical, the quality of the feature maps generally seems an good indicator of final performance: feature maps more resembling to the ground truth at the relatively low resolution (the second row) generally have better performance (cf. the feature maps of NN and A2U).

\begin{table*}[!t]\scriptsize
    \caption{Monocular depth estimation results on NYU Depth V2 with BTS. `HIN' version of IndexNet is used. Best performance is in \textbf{boldface} and second best is \underline{underlined}}
    \centering
    \renewcommand{\arraystretch}{1.2}
    \addtolength{\tabcolsep}{1.1pt}
    \begin{tabular}{@{}l|l|ccc|ccccl@{}}
    \toprule
         & & \multicolumn{3}{c|}{$\tt accuracy$~~$\tt metric\uparrow$} & \multicolumn{5}{c}{$\tt error$~~$\tt metric\downarrow$}  \\
        BTS-ResNet50 \citep{lee2019big} & Params & $\delta < 1.25$ & $\delta < 1.25^2$ & $\delta < 1.25^3$ & Abs Rel & Sq Rel & RMS & RMS (log) & log10\\
        \midrule
        Nearest & 49.5M & 0.865 & 0.975 & 0.993 & 0.119 & 0.075 & 0.419 & 0.152 & 0.051\\
        CARAFE \citep{jiaqi2019carafe} & +0.4M & 0.864 & 0.974 & \underline{0.994} & 0.117 & 0.071 & 0.418 & 0.152 & 0.051\\
        IndexNet \citep{lu2022index} & +44.2M & 0.866 & 0.976 & \textbf{0.995} & 0.117 & 0.071 & 0.416 & 0.151 & \underline{0.050}\\
        A2U \citep{dai2021learning} & +0.2M & 0.860 & 0.973 & 0.993 & 0.121 & 0.077 & 0.429 & 0.156 & 0.052\\
        SAPA \citep{lu2022sapa} & +0.2M & \underline{0.871} & \underline{0.977} & \underline{0.994} & 0.116 & 0.070 & \underline{0.410} & 0.151 & \underline{0.050}\\
        FADE & +2.8M & \textbf{0.875} & \textbf{0.978} & \textbf{0.995} & \textbf{0.114} & \textbf{0.068} & \textbf{0.404} & \textbf{0.147} & \textbf{0.049}\\
        FADE-Lite & +2.5M & 0.870 & \underline{0.977} & \textbf{0.995} & \underline{0.115} & \underline{0.069} & 0.411 & \underline{0.150} & \underline{0.050}\\
    \bottomrule
    \end{tabular}
    \label{tab:depth_estimation}
\end{table*}

\subsection{Monocular Depth Estimation}

Our final task is monocular depth estimation \citep{xian2018monocular}. This task aims to infer the depth from a single image. Compared with other tasks, depth estimation is a mixture of region- and detail-sensitive dense predictions. In a local region, depth values could remain constant (an object plane parallel to the image plane), could be gradually varied (an object plane oblique to the image plane), or could be suddenly changed (on the boundary between different depth planes). The recovery of details in depth estimation is also critical for human perception, because boundary artifacts can be easily perceived by human eyes in many depth-related applications such as 3D ken burns \citep{niklaus20193d} and bokeh rendering \citep{peng2022bokehme}.

\subsubsection{Data Set, Metrics, Baseline, and Protocols}

We use the NYU Depth V2 \citep{silberman2012indoor} dataset and standard depth metrics used by previous work to evaluate the performance, including root mean squared error (RMS) and its log version (RMS (log)), absolute relative error (Abs Rel), squared relative error (Sq Rel), average log$_{10}$ error (log10), and the accuracy with threshold $thr$ ($\delta<thr$). Readers can refer to \citep{lee2019big} for definitions of the metrics. We use BTS\footnote{https://github.com/cleinc/bts} as our baseline and modify all the upsampling stages except for the last one, because there is no guiding feature map at the last stage. We follow the default training setting provided by the authors but set the batch size as $4$ in our experiments (due to limited computational budgets).

\subsubsection{Monocular Depth Estimation Results}

Quantitative and qualitative results are shown in Table~\ref{tab:depth_estimation} and Fig.~\ref{fig:fade_visual}, respectively. Note that FADE requires more number of parameters in this task. The reason is that the number of channels in encoder and decoder features are different, and we need a few $1\times1$ convolutions to adjust the channel number for the gating mechanism. Overall, FADE reports consistently better performance in all metrics than other competitors, and FADE-Lite is also the steady second best. It is worth noting that A2U degrades the performance, which suggests only improving detail delineation is not sufficient for depth estimation. FADE, however, fuses the benefits of both detail- and region-aware upsampling capable of simultaneous detail delineation and regional preservation. We believe this is the reason why FADE behaves remarkably on this task.


\begin{table*}[!t]
    \caption{Ablation study on the source of features, the way for upsampling kernel generation, and the effect of the gating mechanism. `en' is for encoder, and `de' for decoder. Best performance is in \textbf{boldface}}
    \label{tab:ablation_study}
    \centering
    \renewcommand{\arraystretch}{1.2} 
    \addtolength{\tabcolsep}{5pt}
    \begin{tabular}{@{}lccccc|cccc@{}}
        \toprule
        No. & \multicolumn{3}{c}{SegFormer / A2U Matting} & \multicolumn{2}{c|}{$\tt Segm$ -- $\tt accuracy\uparrow$} & \multicolumn{4}{c}{$\tt Matting$ -- $\tt error\downarrow$}\\
        & source of feat. & kernel gen. & fusion & mIoU & bIoU & SAD & MSE & Grad & Conn\\
        \midrule
        b1 & en &  & & 42.75 & 31.00 & 34.22 & 0.0087 & 15.90 & 32.03\\
        b2 & de &  & & 42.82 & 29.84 & 41.01 & 0.0118 & 21.39 & 39.01 \\
        b3 & en \& de & naive & & 43.27 & 31.55 & 32.41 & 0.0083 & 16.56 & 29.82\\
        b4 & en \& de & semi-shift & & 43.33 & 32.06 & 31.78 & 0.0075 & 15.12 & 28.95\\
        b5 & en \& de & semi-shift & skipping & 43.22 & 31.85 & 32.64 & 0.0076 & 15.90 & 29.92\\
        b6 & en \& de & semi-shift & gating & \textbf{44.41} & \textbf{32.65} & \textbf{31.10} & \textbf{0.0073} & \textbf{14.52} & \textbf{28.11}\\
        \bottomrule
    \end{tabular}
\end{table*}

\subsection{Ablation Study}

Here we conduct ablation studies to justify our three design choices. We follow the settings in segmentation and matting, because they are sufficiently representative to indicate region- and detail-sensitive tasks. In particular, we explore how performance is affected by the source of features, the way for upsampling kernel generation, and the use of the gating mechanism. We build six baselines:
\begin{enumerate}

    \item[1)] b1: \textit{encoder-only}. Only encoder features go through $1\times1$ convolution for channel compression ($64$ channels), followed by $3\times3$ convolution layer for kernel generation;
    
    \item[2)] b2: \textit{decoder-only}. This is the CARAFE baseline \citep{jiaqi2019carafe}. Only decoder features go through the $1\times1$ and $3\times3$ convolution for kernel generation, followed by Pixel Shuffle; 
    
    \item[3)] b3: \textit{encoder-decoder-naive}. NN-interpolated decoder features are first concatenated with encoder features, and then the same two convolutional layers are applied;
    
    \item[4)] b4: \textit{encoder-decoder-semi-shift}. Instead of using NN interpolation and standard convolutional layers, we use semi-shift convolution to generate kernels as in FADE;
    
    \item[5)] b5: b4 with \textit{skipping}. We directly skip the encoder features as in feature pyramid networks \citep{lin2017feature}; 
    
    \item[6)] b6: b4 with \textit{gating}. The full implementation of FADE.
    
\end{enumerate} 
Results are shown in Table \ref{tab:ablation_study}. By comparing b1, b2, and b3, the results confirm the importance of both encoder and decoder features for upsampling kernel generation. By comparing b3 and b4, semi-shift convolution is superior than naive implementation in the way of generating upsampling kernels. As aforementioned, the rationale behind such a superiority can boil down to the granular control 
on the per-point contribution in the kernel (Section \ref{sec:fade}). We also note that, even without gating, the performance of FADE already surpasses other upsampling operators (b4 vs.\ Table~\ref{tab:performance}), which means the task-agnostic property is mainly due to the joint use of encoder and decoder features and the semi-shift convolution. In addition, skipping in these two task is clearly not the optimal way to move encoder details to decoder features, at least worse than the gating mechanism (b5 vs.\ b6). Hence, we think gating is generally beneficial.

\subsection{Limitations and Further Discussions}
\label{subsec:limitation}

\paragraph{Computational Overhead.} Despite FADE outperforms CARAFE in $4$ out of $6$ tasks, FADE processes $5$ times data more than CARAFE and thus consumes more FLOPs due to the involvement of high-res encoder features. Our efficient implementations do not change this fact but only help prevent extra calculations on interpolated decoder features. A thorough comparison of the computational complexity and inference time of different dynamic upsampling operators can be found in Appendix~\ref{asec:complexity}.

\paragraph{Prerequisite of Using FADE.} The use of the gating mechanism in FADE requires an equal number of channels of encoder and decoder features. Therefore, if the channel number differs, one needs to add a $1\times1$ convolution layer to align the channel number. However, this would introduce additional parameters, for example depth estimation with BTS. If the gate is not used, \textit{i.e.}, FADE (G=1), this trouble could be saved. In addition, if there is no high-res feature guidance, for instance, the last upsampling stage in BTS or in image super-resolution tasks, FADE cannot be applied as well.

\paragraph{When to Use the Gating Mechanism.} At our initial design \citep{lu2022fade}, we mainly consider the one-class-one-value mapping as in semantic segmentation or regressing a dense 2D map as in image matting, but do not explore instance-level tasks like object detection and instance segmentation, where the situation differs from what we initially claim. We find that the high-res encoder feature plays an important role in localization. If forcing the feature map to be alike to that in semantic segmentation, the model cannot learn instance-aware information effectively. In this case the gating mechanism can fail, and we propose to use direct addition ($G=1$) as a substitution. One should also be aware that, semi-shift convolution can introduce encoder noise in the generated kernel such that the precise localization of bounding box could be affected (the obviously lower $AP_{75}$ of FADE than CARAFE in object detection and instance segmentation).

\paragraph{General Value of Upsampling to Dense Prediction.} As closing remarks, here we tend to share our insights on the general value of upsampling to dense prediction. Compared with other operators or modules studied in dense prediction models, upsampling operators have received less attention. While we have conducted extensive experiments to demonstrate the effectiveness of upsampling, one may still raise the question: Is upsampling an intrinsic factor to influence the dense prediction performance? Indeed current mainstream ideas are to scale the model \citep{tan2019efficientnet,zhai2022scaling}, and results from Table~\ref{tab:segformer_scaling} also indicate that, under a certain evaluation metric, a strong backbone with a simple bilinear upsampling is sufficient. Yet, we remark that, if one keep pursuing the increment of a certain metric in a specific task, \textit{e.g.}, mIoU in semantic segmentation, some other important things would be overlooked such as the boundary quality. From also Table~\ref{tab:segformer_scaling}, we can observe that enhanced upsampling steadily boosts the bIoU metric. This is only in segmentation.
From a broad view across different dense prediction tasks, the value of upsampling can even be greater, particularly for low-level tasks. For instance, it has been reported that, with learned upsampling, the Deep Image Prior model can use $95\%$ fewer parameters to achieve superior denoising results than existing methods \citep{liu2023devil}. Our previous experience in matting also suggests inappropriate upsampling even cannot produce a reasonable alpha prediction \citep{lu2022index}. From the perspective of architecture design, different operators or modules function differently, but their ultimate goal is alike, \textit{i.e.}, learning high-quality features. If enabling an upsampling operator that has a high probability of being used in an encoder-decoder architecture to have equivalent or even better functions implemented by other optional modules, the architecture design could be simplified. Task-agnostic upsampling at least demonstrates such a potential. Indeed upsampling matters. We believe the value of upsampling is not only about improved performance but also about the design of new, effective, efficient, and generic encoder-decoder architectures.

Another closely-related question is that: Does one still need new fundamental (upsampling) operators, particularly in the era of vision foundation models \citep{caron2021emerging,radford2021learning,rombach2022high,Kirillov_2023_ICCV} when the idea of scaling typically wins? Indeed current foundation models are made of standard operators such as convolutional layers \citep{radford2021learning} and self-attention blocks \citep{caron2021emerging}. The classic U-Net architecture \citep{ronneberger2015u} is also used in StableDiffusion \citep{rombach2022high}. The adoption of sophisticated operators or architectures seem unnecessary if the model capacity reaches to a certain level. Yet, we note a phenomenon that the SAM model \citep{Kirillov_2023_ICCV} still cannot generate accurate mask boundaries. We believe one of the reasons is that it still uses the deconvolution upsampling in the decoder, which smoothes boundaries. Hence, we think designing fundamental and task-agnostic network operators would remain to be an active research area. Here we make a tentative prediction: a real sense of the vision foundation model should be made of task-agnostic operators. We expect this work can inspire the new design of such operators.

\section{Conclusion}

In this paper, we provide feature upsampling with three levels of meanings: i) being basic, the ability to increase spatial resolution; ii) being effective, the capability of improving performance; and iii) being task-agnostic, the generality across tasks. In particular, to achieve the third property, we propose FADE, a novel, plug-and-play, and task-agnostic upsampling operator by fully fusing the assets of encoder and decoder features. For the first time, FADE demonstrates that task-agnostic upsampling is made possible across both region- and detail-sensitive dense prediction tasks, outperforming or at least being comparable with the previous best upsampling operators. We explain the rationale of our design with step-to-step analyses and also share our view points from considering what makes for generic feature upsampling. Our core insight is that an upsampling operator should be able to dynamically trade off between detail delineation and semantic preservation in a content-aware manner.

We encourage others to try this operator on many more dense prediction tasks, particularly on low-level tasks such as image restoration.
So far, FADE is designed to maintain the simplicity by only implementing linear upsampling, which leaves ample room for further improvement, \textit{e.g.}, by exploring additional nonlinearity. 

\paragraph{Funding} This work is supported by the National Natural Science Foundation of China Under Grant No. 62106080 and the Hubei Provincial Natural Science Foundation of China under Grant No. 2024AFB566.

\appendices

\renewcommand{\thetable}{\thesection\arabic{table}}

\section{Comparison of Computational Complexity}
\label{asec:complexity}

\begin{table}[!t] \tiny
    \caption{Computational complexity and parameters of FADE and other upsampling operators. C: number of channels of encoder and decoder features; d: number of compressed channels; K: upsampling kernel size; H, W: height and width of the \textbf{decoder} feature map. G=1 indicates no gating mechanism}
    \label{tab:complexity}
    \centering
    \renewcommand{\arraystretch}{1.1}
    \addtolength{\tabcolsep}{-1.5pt}
    \begin{tabular}{@{}llcc@{}}
    \toprule
        Module & Operation & FLOPs ($\times HW$) & Params\\
        \midrule
        CARAFE & Kernel generation & $Cd\!+\!36K^2d$ & $Cd\!+\!36K^2d$ \\
         & Feature assembly & $4K^2C$ & $0$ \\
         & \textbf{Total} & $Cd\!+\!36K^2d\!+\!4K^2C$ & $Cd\!+\!36K^2d$ \\
        \midrule
        IndexNet & Kernel generation & $32C^2\!+\!8C$ & $32C^2\!+\!8C$ \\
        HIN & Feature assembly & $4C$ & $0$ \\
         & \textbf{Total} & $32C^2\!+\!12C$ & $32C^2\!+\!8C$ \\
        \midrule
        IndexNet & Kernel generation & $68C^2$ & $68C^2$ \\
        M2O & Feature assembly & $4C$ & $0$ \\
         & \textbf{Total} & $68C^2\!+4C$ & $68C^2$ \\
        \midrule
        A2U & Kernel generation & $73C\!+\!4K^2\!$ & $4K^2C\!+\!2C$ \\
         & Feature assembly & $4K^2C$ & $0$ \\
         & \textbf{Total} & $73C\!+\!4K^2\!+\!4K^2C$ & $4K^2C\!+\!2C$ \\
        \midrule
        SAPA & Kernel generation & $5Cd+4K^2d$ & $2Cd$ \\
         & Feature assembly & $4K^2C$ & $0$ \\
         & \textbf{Total} & $5Cd+4K^2d+4K^2C$ & $2Cd$ \\
        \midrule
        FADE & Kernel generation & $5Cd\!+\!45K^2d$ & $2Cd\!+\!9K^2d$ \\
         & Feature assembly & $4K^2C$ & $0$ \\
         & Gated fusion & $9C$ & $C$\\
         & \textbf{Total} & $5Cd\!+\!4K^2C\!+\!45K^2d\!+\!9C$ & $2Cd\!+\!9K^2d\!+\!C$ \\
         & \textbf{Total} (G=1) & $5Cd\!+\!45K^2d\!+\!4K^2C$ & $2Cd\!+\!9K^2d$ \\
        \midrule
        FADE & Kernel generation & $5CK^2\!+\!45K^2$ & $2CK^2\!+\!9K^2$ \\
        Lite & Feature assembly & $4K^2C$ & $0$ \\
         & Gated fusion & $9C$ & $C$\\
         & \textbf{Total} & $5CK^2\!+\!4K^2C\!+\!45K^2\!+\!9C$ & $2CK^2\!+\!9K^2\!+\!C$ \\
         & \textbf{Total} (G=1) & $5CK^2\!+\!45K^2\!+\!4K^2C$ & $2CK^2\!+\!9K^2$ \\
    \bottomrule
    \end{tabular}
\end{table}

\begin{table}[!t] \scriptsize
    \caption{Comparison of inference time among different upsampling operators. Time is tested on a single Nvidia GTX 3090 GPU on a server with Intel Xeon Gold 6226R @ 2.90 GHz CPUs}
    \label{tab:latency}
    \centering
    \renewcommand{\arraystretch}{1.1}
    \addtolength{\tabcolsep}{-3.5pt}
    \begin{tabular}{@{}cccccccccc@{}}
    \toprule
        Upsampler & Bilinear & CARAFE & IndexNet & A2U & FADE & FADE-Lite \\
        \midrule
        Time (ms) & 1.5 & 11.1 & 26.4 & 24.2 & 20.2 & 17.6\\
    \bottomrule
    \end{tabular}
\end{table}

A favorable upsampling operator, being part of overall network architecture, should not significantly increase the computation cost. This issue is not well addressed in IndexNet as it introduces 
many parameters and much computational overhead \citep{lu2019indices}. In this part we analyze the computational workload and memory occupation among different dynamic upsampling operators. We first compare the FLOPs and number of parameters in Table~\ref{tab:complexity}. FADE requires more FLOPs than CARAFE (note that FADE processes $5$ times more feature data than CARAFE), but less parameters when the number of channels is small. For example, when $C=256$, $d=64$, $K=5$, and $H=W=112$, CARAFE and FADE cost $2.50$ and $4.56$ GFLOPs, respectively; the number of parameters are $74$ K and $47$ K, respectively. FADE-Lite, in the same setting, costs only $1.53$ GFLOPs and $13$ K parameters. In addition, we also test the inference speed by upsampling a random feature map of size $256\times 120 \times 120$ (a guiding map of size $256\times 240\time 240$ is used if required). The inference time is shown in Table~\ref{tab:latency}. Among compared dynamic upsampling operators, 
FADE and FADE-Lite are relatively efficient given that they process five times more data than CARAFE. 
We also test the practical memory occupation of FADE on SegFormer-B1 \citep{xie2021segformer}, with $6$ upsampling stages. Under the default training setting, SegFormer-B1 with bilinear upsampling costs $22,157$ MB GPU memory. With the H2L implementation of FADE, it consumes $24,879$ MB, $2,722$ MB more than the original one. The L2H one reduces the memory cost by $24.2\%$ (from $2,722$ to $2,064$ MB), and is within an acceptable range compared with the decoder-only upsampling operator CARAFE ($664$ MB) if taking the five times more data into account.

\bibliographystyle{spbasic}
\bibliography{egbib}

\end{document}